\documentclass[10pt,twocolumn,letterpaper]{article}

\usepackage{iccv}
\usepackage{times}
\usepackage{epsfig}
\usepackage{graphicx}
\usepackage{amsmath}
\usepackage{amssymb}
\usepackage{microtype}
\usepackage[table,xcdraw]{xcolor}
\usepackage{caption}
\usepackage{float}
\usepackage{placeins}
\usepackage{color, colortbl}
\usepackage{stfloats}
\usepackage{enumitem}
\usepackage{tabularx}
\usepackage{xstring}
\usepackage{multirow}
\usepackage{xspace}
\usepackage{url}
\usepackage{subcaption}
\usepackage{xcolor}
\usepackage{algorithm}
\usepackage{algorithmic}
\usepackage{setspace}
\usepackage{bm}
\usepackage{comment}
\usepackage{cases}
\usepackage[switch]{lineno}
\usepackage[hang,flushmargin]{footmisc}
\usepackage[normalem]{ulem}
\usepackage{booktabs}
\usepackage[accsupp]{axessibility}

\usepackage[breaklinks=true,bookmarks=false]{hyperref}

\iccvfinalcopy 

\def\iccvPaperID{4386} 

\ificcvfinal\fi

\begin{document}

\title{Knowledge Restore and Transfer for Multi-Label Class-Incremental Learning}

\author{Songlin Dong\textsuperscript{\rm 1 \#}, Haoyu Luo\textsuperscript{\rm 1 \#}, Yuhang He\textsuperscript{\rm 1 }{\thanks{Yuhang He is the corresponding author; \# Songlin Dong and Haoyu Luo are co-first authors}} , Xing Wei\textsuperscript{\rm 2}, Jie Cheng\textsuperscript{\rm 3}, Yihong Gong\textsuperscript{\rm 2}\\
\textsuperscript{\rm 1}College of Artificial Intelligence, Xi’an Jiaotong University\\
\textsuperscript{\rm 2}School of Software Engineering, Xi'an Jiaotong University\\
\textsuperscript{\rm 3}ACS Lab, Huawei Technologies, Shenzhen, China\\
{\tt\small \{dsl972731417,luohaoyu,hyh1379478\}@stu.xjtu.edu.cn}  \\
{\tt\small chengjie8@huawei.com}, {\tt\small \{weixing,ygong\}@mail.xjtu.edu.cn}}

\maketitle
\ificcvfinal\thispagestyle{empty}\fi


\begin{abstract}
\vspace{-0.1cm}
Current class-incremental learning research mainly focuses on single-label classification tasks while multi-label class-incremental learning (MLCIL) with more practical application scenarios is rarely studied. Although there have been many anti-forgetting methods to solve the problem of catastrophic forgetting in single-label class-incremental learning, these methods have difficulty in solving the MLCIL problem due to label absence and information dilution problems. To solve these problems, we propose a Knowledge Restore and Transfer (KRT) framework containing two key components. First, a dynamic pseudo-label (DPL) module is proposed to solve the label absence problem by restoring the knowledge of old classes to the new data. Second, an incremental cross-attention (ICA) module is designed to maintain and transfer the old knowledge to solve the information dilution problem. Comprehensive experimental results on MS-COCO and PASCAL VOC datasets demonstrate the effectiveness of our method for improving recognition performance and mitigating forgetting on multi-label class-incremental learning tasks. The source code is available at~\href{https://github.com/witdsl/KRT-MLCIL}{https://github.com/witdsl/KRT-MLCIL}.
\end{abstract}

\vspace{-0.2cm}
\section{Introduction}
\label{sec:intro}
\begin{figure}[t]
\setlength{\abovecaptionskip}{-0.25cm} 
\setlength{\belowcaptionskip}{-0.55cm} 
\begin{center}
    \includegraphics[width=0.50\textwidth]{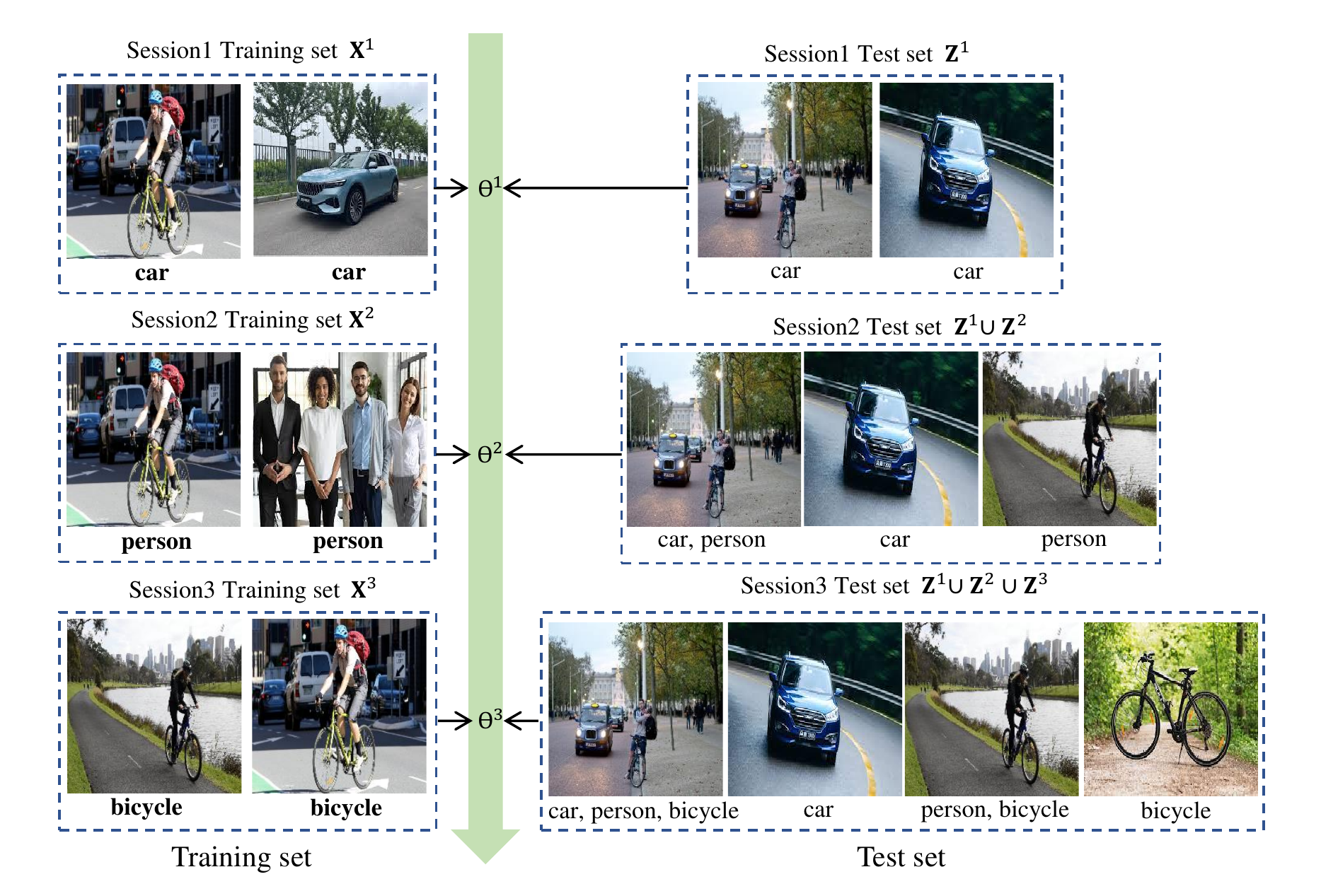}
\end{center}
\caption{The illustration of the multi-label class-incremental learning task. Supposed there are three categories in total: car, person, and bicycle, which are incrementally learned in three sessions. ($\Theta^1$,$\Theta^2$,$\Theta^3$ are the models that are continuously trained)}
\label{fig:mi}
\end{figure}

\vspace{-0.1cm}
Class-Incremental Learning (CIL)~\cite{der+,dong2021few,iod2022,seg_il2022,iod2017} aims to continuously learn new classes as well as maintain the performance of old classes. When applied to a classification task, most existing CIL methods~\cite{topic,podnet,der+,l2p} generally first assume each image only contains a single object, and then develop anti-forgetting mechanisms to learn new classes without forgetting the old ones, \ie, the single-label CIL problem. In real-world applications, however, an image usually contains multiple objects (\eg, a \textbf{man} riding his \textbf{bicycle}) and the provided labels are often \emph{category-incomplete} due to incrementally defined classes. For example, as shown in Figure~\ref{fig:mi}, only a class label `car' is provided in session 1. Then, newly defined `person' and `bicycle' are annotated on previously and newly collected images in sessions 2 and 3, respectively. The classification models are expected to recognize all the newly and previously encountered categories (\eg, recognize `car' in session 1 and recognize `car', `person' and `bicycle' in session 3). Taking category-incomplete labels as inputs in different sessions and having the capacity of recognizing all the encountered categories, we term this ability as \textit{multi-label class-incremental learning} (MLCIL). This is a more challenging but practical problem for real-world applications.



A simple solution of the MLCIL problem is to fine-tune a multi-label classification (MLC) model using the training samples of each new session. However, this method leads to catastrophic forgetting~\cite{mccloskey1989catastrophic}, where classification accuracy on old classes deteriorate drastically. Another feasible solution is introducing single-label CIL methods~\cite{ICARL,podnet,der+,topic} to the MLCIL problem by adopting ML classification head~\cite{asl2020}. 
Most existing CIL methods~\cite{ICARL,podnet,der+,tao2020topology} solve the catastrophic forgetting problem through replaying a portion of representative old exemplars (ER) and designing different knowledge distillation (KD) losses~\cite{ICARL,der+,podnet} to transfer knowledge from old sessions to new sessions.

However, adapting these methods to the MLCIL problem is faced with two major challenges: 1) the \emph{label absence} of old classes. At each session, images are only annotated with new classes even if they contain old class objects. The absence of old class labels makes these images negative samples of the old classes, thus leading to more serious catastrophic forgetting. For example, in the training session 3 in Figure~\ref{fig:mi}, the right image contains a \textbf{person} riding his \textbf{bicycle} in front of a \textbf{car} is labeled as `\textbf{bicycle}'. Training this image with the single label `\textbf{bicycle}' makes it a negative sample of \textbf{car} and \textbf{person}, leading to catastrophic forgetting of these two old classes. (2) The \textit{information dilution} during knowledge transfer. To alleviate forgetting, most existing CIL methods retain an old-sample buffer and transfer the knowledge of old samples to new sessions by knowledge distillation. However, retaining data is often not allowed in practice due to privacy and safety concerns. Even worse, the widely used KD techniques will omit detailed information~\cite{der}. This makes the MLC classifier have less ability to recognize the challenging (\eg, small or occluded) objects especially when there are multiple objects and dramatically decreases the multi-label classification performance.

To address the above challenges, in this paper, we propose a \textit{knowledge restore and transfer}~(KRT) framework for the MLCIL problem. The KRT framework contains two major modules: 1) a dynamic pseudo-label (DPL) module to restore the knowledge of old classes and 2) an incremental cross-attention (ICA) module to transfer knowledge across different sessions. More specifically, in the DPL module, we feed new data to the old model and generate pseudo labels of the old classes according to dynamic thresholds. The pseudo labels restore the knowledge of old classes and are combined with the current labels (of new classes) to jointly train the new model. In the ICA module, a unified \textbf{knowledge transfer}~(KT) token for all the sessions and multiple \textbf{knowledge retention}~(KR) tokens for different sessions, respectively, are designed to maintain and transfer the old knowledge. The KT token aims to learn knowledge-transfer-related information and is continuously trained across all the sessions. The KR token aims to learn category-related knowledge of the current session and is only trained on the current session. By incorporating the KT token with an old session KR token, the ICA outputs a session-specific embedding for the input image, which transfers the knowledge of old session to the current session. During the training and inference process of the $t$-th session, the ICA outputs a total number of $t$ session-specific embeddings using the current and $t-1$ previously obtained KR tokens, and leverage these embeddings for multi-label classification. Preserving the old session knowledge to the KR token and transferring them to the current session using the KT token, the ICA module can effectively preserve and transfer knowledge for incremental learning and solve the problem of information dilution caused by KD. On this basis, a token loss is designed to optimize the ICA module to transfer knowledge and prevent the forgetting of old knowledge. 

For extensive evaluation, we construct the MLCIL baselines by adapting the latest multi-label methods~\cite{asl2020,PRS,OCDM} and state-of-the-art CIL methods~\cite{LWF,oewc,ICARL,podnet,ERbase,BIC,tao2020topology,der+,l2p} to this new problem and comparing our KRT with them. We conduct comprehensive experiments on popular MLC datasets, including MS-COCO~\cite{coco2014}, and PASCAL VOC~\cite{voc2007}. To summarize, our main contributions include:
\begin{itemize}
\item We propose a knowledge restore and transfer~(KRT) framework, which is one of the first attempts, to address the multi-label class-incremental learning~(MLCIL) problem.
\item We design a dynamic pseudo-label (DPL) module to solve the label absence problem by restoring the knowledge of old classes to the new data.
\item We develop an incremental cross-attention~(ICA) module with session-specific KR tokens storing knowledge and a unified KT token transferring knowledge to solve the information dilution problem.
\item Extensive experiments on MS-COCO and PASCAL VOC demonstrate that the proposed method achieves state-of-the-art performance on the MLCIL task.
\end{itemize}

\vspace{-0.15cm}
\section{Related Work}
\label{sec:rela}
\vspace{-0.05cm}
\subsection{Single-label Incremental Learning}
\vspace{-0.05cm}
\textbf{Regularization-based} methods introduce a regularization term in the loss function so that the updated parameter retains old knowledge. 1) Parameter regularization: reduce the variation of parameters related to old tasks~\cite{EWC,SI,MAS,oewc}. EWC~\cite{EWC} uses a fisher matrix to preserve the important parameters of the historical tasks. Then oEWC~\cite{oewc} and other methods~\cite{SI,REWC,MAS} are constantly improving the parameter importance calculation. 2) Data regularization: consolidate the old knowledge by using previous models as soft teachers while learning the new data~\cite{LWF,lwm}. For example, LWF~\cite{LWF} exploits knowledge distillation to mitigate forgetting. 

\textbf{Rehearsal-based} methods store a set of exemplars as representative of the old data to train with new data from the current task. Most of the rehearsal-based methods are used to solve class-incremental learning~\cite{ICARL,tao2020topology,co2l,podnet,der+} problem. Early rehearsal method ER~\cite{ERbase} simply constructs a memory buffer to save samples from old tasks to retrain with new data. On this basis, DER++~\cite{der+} proposes knowledge distillation penalties on data stored in the memory buffer. Moreover, the iCaRL~\cite{ICARL} and its variants~\cite{EEIL,lucir,podnet,kang2022class} prevent forgetting by selecting exemplars by using the herding~\cite{herding} technique and designing different distillation losses. BIC~\cite{BIC} and other methods~\cite{bic2020,il2m} perform an additional bias correction process to modify the classification layer. TPCIL~\cite{tao2020topology} constructs an EHG to model the feature space and propose a topology-preserving loss to maintain the feature space topology. Recent methods~\cite{aanet,CwD,ashok2022class} propose adaptive aggregation networks or mimic the feature space distribution of oracle to improve the above rehearsal-based methods. 

\textbf{Architectural-based} methods provide independent parameters for each task to prevent possible forgetting. Most of the architectural methods require additional task oracle and are restricted to the multi-head setup (Task-IL scenario). Abati et al.~\cite{PackNet,hat,CCGN} propose different strategies to isolate the old and new task parameters and Rusu et al.~\cite{PNN} replicates a new network for each task to transfer prior knowledge through lateral connections to new tasks. The latest architectural methods~\cite{der,dytox,foster} combined with the rehearsal methods achieve a better anti-forgetting effect. These methods dynamically expand or prune the network parameters to accommodate the new data at the expense of limited scalability. Moreover, L2P~\cite{l2p} exploits dependent prompting methods based on a pre-trained ViT model for continual learning, which achieves state-of-the-art results on multiple single-label incremental learning tasks. 
\vspace{-0.1cm}
\subsection{Multi-label Incremental Learning}
\vspace{-0.05cm}
\textbf{Multi-label classification} aims to gain a comprehensive understanding of objects and concepts in an image and the proposed methods can be categorized into two main directions: label dependency~\cite{MCRN2013,RMAM2017,chen2021learning,C-Trans,q2l} and loss function~\cite{2020disloss,focal2017,asl2020}. In this paper, we adopt asymmetric loss~(ASL)~\cite{asl2020} as the classification loss to achieve our KRT and all other compared methods. 

\textbf{Multi-label online incremental learning.} Online incremental learning~\cite{GEM} involves organizing tasks into a non-stationary data stream, where the agent can only receive a mini-batch of task samples from the data stream and traverse the data of each task only once. Currently, not only has there been a large amount of works~\cite{yoon2021online,shim2021online,sun2022information,guo2020online,han2023online} on single-label OIL, but multi-label online incremental learning has also gradually received widespread attention. For example, Du et al.~\cite{AGCN} construct the relationship between labels and design a graph convolutional network to learn them. PRS~\cite{PRS} proposes sample-in/sample-out mechanisms to balance the class distribution in memory. Furthermore, OCDM~\cite{OCDM} proposes a greedy algorithm to control the class distribution in memory fast and efficiently when the data stream consists of multi-label samples. 

\section{Method}
\label{sec:method}

\begin{figure*}[t]
\setlength{\abovecaptionskip}{-0.2cm} 
\setlength{\belowcaptionskip}{-0.4cm} 
\begin{center}
    \includegraphics[width=0.96\textwidth,height=0.43\textwidth]{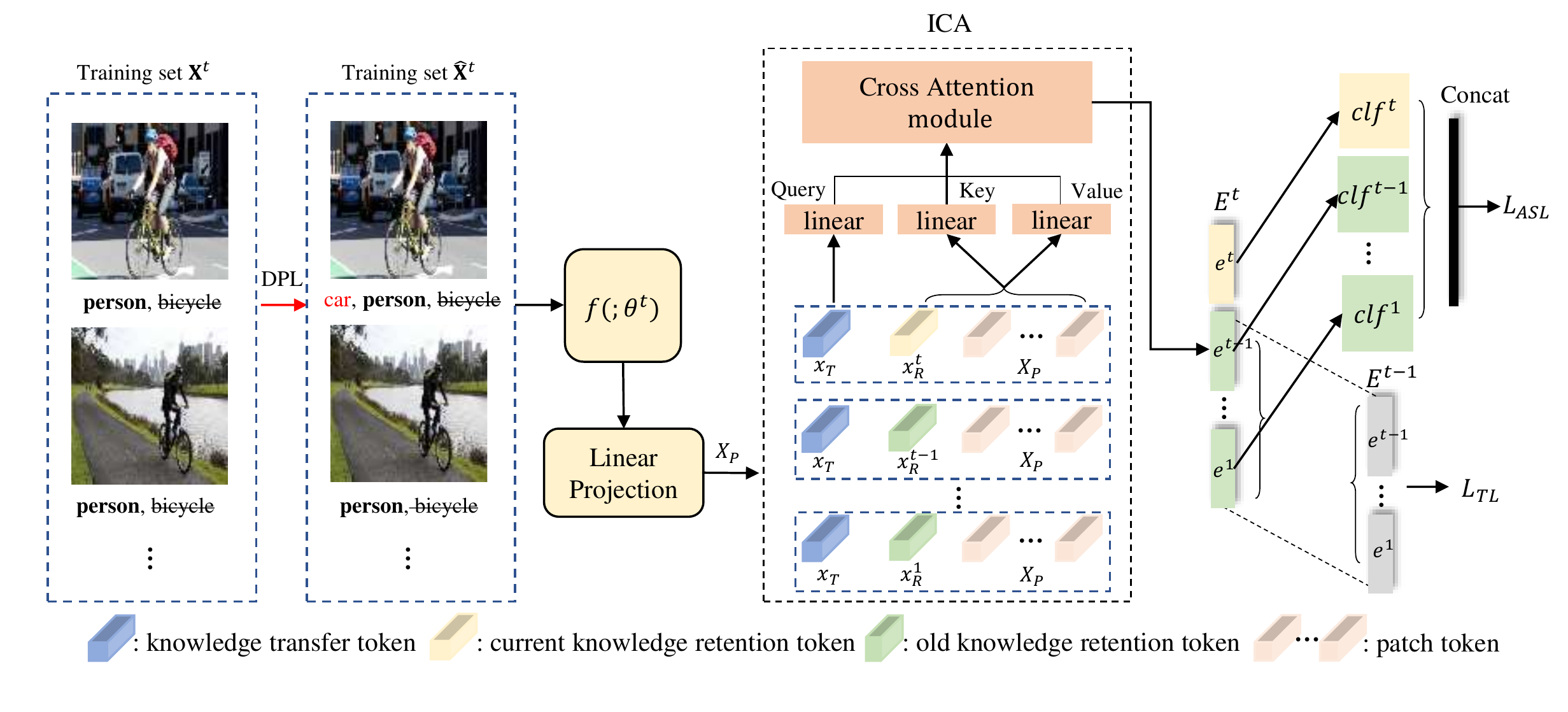}
\end{center}
\caption{The framework of our proposed KRT for MLCIL problem. The image is first to restore old knowledge by dynamic pseudo-label~(DPL) module and obtains the pseudo label `\textcolor{red}{car}' (`\sout{bicycle}' indicates that the class has not been defined yet). Then, we feed the restored image into the feature extractor $f(;\theta^t)$ and the linear projection to obtain patch token $X_P$. Finally we fed the $X_P$ into incremental cross-attention~(ICA) module to acquire final output logits ($E^{t-1}$ and $E^{t}$ are outputs of previous and current ICA module).}
\label{fig:framework}
\end{figure*}

\subsection{Problem Formulation}
\label{sec:protocol}
\vspace{-0.1cm}
Assuming that there are a total number of $T$ incremental sessions $ \left\{\mathbf{D}^1,\mathbf{D}^2,...,\mathbf{D}^T \right\}$, where $\mathbf{D}^t=\left\{\mathbf{X}^t, \mathbf{Z}^t\right\}$ is consisted of a training set $\mathbf{X}^t$ and a test set $\mathbf{Z}^t$. Each training set is defined as $\mathbf{X}^t=\left \{ \left ( \mathbf{x}_{i}^{t}, \mathbf{y}_{i}^{t} \right ) \right \}$, where $\mathbf{x}_{i}^{t}$ is the $i$-th training sample and $\mathbf{y}^{t}_{i}\subseteq\mathbf{C}^t$ is a label set with $1\leq |\mathbf{y}^{t}_{i}| \leq |\mathbf{C}^t|$. $\mathbf{C}^t$ denotes the class collection at the $t$-th session and $\forall m,n~(m\neq n),~\mathbf{C}^{m} \cap\mathbf{C}^{n}=\varnothing$. With the MLCIL setting, a \textit{unified} \textbf{multi-label classification} model will be incrementally trained across the $T$ sessions. At each session $t$, only $\mathbf{X}^{t}$ is available during training and the model is evaluated on a combination of test sets $\mathbf{Z}^{1\sim t}=\mathbf{Z}^1\cup\cdots\cup\mathbf{Z}^t$ and is expected to recognize all the encountered classes $\mathbf{C}^{1\sim t}=\mathbf{C}^1\cup\cdots\cup\mathbf{C}^t$.
\vspace{-0.1cm}
\subsection{Framework}
\label{sec:framework}
\vspace{-0.1cm}
Given a multi-label classifier composing of a feature extractor $f(;\theta)$ and a classification head $\varphi(;\phi)$. We use $\Theta=\left\{\theta,\phi \right\}$ to denote the total parameters. First, we train a base model $\Theta^{1}$ using $\mathbf{X}^{1}$ with the ASL loss~\cite{asl2020}. Then, we incrementally fine-tune the base model using $\mathbf{X}^{2},...,\mathbf{X}^{T}$, and get $\Theta^{2},...,\Theta^T$. At the session $t$ $(t>1)$, the classification head is expanded for new classes by adding $N^{t}=|\mathbf{C}^t|$ output neurons.

Figure~\ref{fig:framework} illustrates the framework of the proposed KRT, which composed of two core designs: DPL and ICA modules. At each session $t$, we first feed the training set $\mathbf{X}^t$ to the DPL module to generate pseudo labels of the old classes, which combine with the current labels as the new input $\hat{\mathbf{X}}^t$ to jointly train the new model $\Theta^t$. Second, the patch tokens $X_P$ are fed to ICA module to transfer knowledge across different sessions. The $x_{T}$ (blue) is knowledge transfer~(KT) token which is trained across all the sessions. The $x^t_R$ (yellow) and $x_R^{1\sim t-1}$ (green) are knowledge retention~(KR) tokens of current and old sessions, respectively. By incorporating the KT token with current and old KR tokens, the ICA output a total number of $t$ session-specific embeddings~($e^{1\sim t}$) for current session multi-label classification. Finally, a token loss $\mathbf{L}_{TL}$ is designed to jointly optimize the ICA module. The following part of this section provides detailed descriptions of these two components.

\vspace{-0.1cm}
\subsection{Dynamic Pseudo-Label Module}
\vspace{-0.1cm}
\label{dpl}
To prevent the catastrophic forgetting caused by the \textit{label absence} problem, we propose a dynamic pseudo-label~(DPL) module. Concretely, in the session $t$, given an input image $\mathbf{x}^t_{i}$ and a previous model $\Theta^{t-1}$, where $\Theta^{t-1}$ has already learned the knowledge of $K$ classes. We utilize the model $\Theta^{t-1}$ to perform an inference on $\mathbf{x}^t_i$ and get the classification probabilities $\mathbf{p_i}=\{p_1,p_2,...,p_{K}\}$ of $K$ classes, where $p_k\in(0,1)$ denotes the probability of class $k$. If $p_k\geq \eta$, the image is very likely to contain the $k$-th category object, and $p_k < \eta$ denotes the opposite. Here, $\eta \in (0,1)$ is the initial threshold. Finally, we compose pseudo-label set $\mathbf{S}^t$ of all pseudo labels generated by training samples in the current session. 

In general, we employ model $\Theta^{t-1}$ to infer the training set $\mathbf{X}^{t}$ in this session and merge the obtained old pseudo-label set $\mathbf{S}^t$ with the label set $\mathbf{Y}^t$ of this session as new ground truth $\hat{\mathbf{Y}}^t$. Then we use the updated training set $\hat{\mathbf{X}}^{t}$ to train the model $\Theta^{t}$. 
However, as the number of learning sessions increases, the abundance of pseudo labels can impede the ability to acquire knowledge about new classes. Additionally, the model inevitably forgets the knowledge of old classes, particularly those learned early on, resulting in the generation of inaccurate pseudo labels. To address these issues, we propose a straightforward yet efficient method, called dynamic threshold adjustment, which reduces the incidence of falsely generated pseudo labels during the incremental process. Specifically, before each incremental session begins, we dynamically adjust the threshold $\eta^t$ based on $\beta^t$ and $\mu^t$. The $\beta^t=\frac{|\mathbf{S}^t|}{M^t}$ is the number of average pseudo labels per image at the current session, where $|\mathbf{S}^t|$ and $M^t$ are the number of generated pseudo labels and training samples at the current session, respectively. The $ \mu^{t}=(\frac{|\textbf{C}^t_o|}{|\mathbf{C}_{a}|} \times \mu$) is the target value, where $|\textbf{C}^t_o|$ is the number of old classes that have been learned, $|\mathbf{C}_{a}|$ is the total number of classes in the dataset and the $\mu$ is a hyper-parameter. The detailed of DPL algorithm is written in \textbf{Appendix}. The ablation study~\ref{ablation} have proven that the DPL module adapts well to the multi-label incremental learning task, effectively restores old knowledge, and solves the catastrophic forgetting problem.

\subsection{Incremental Cross-Attention Module}
\vspace{-0.1cm}
\subsubsection{Build the ICA Module}
\vspace{-0.1cm}
\
In this section, we introduce the construction of the incremental cross-attention module. Concretely, for each input image, the output of the feature extractor $f(;\theta)$ is $\mathcal{F}\in \mathbb{R}^{h \times w \times c}$, where $h,w$ denote the height and width of feature map respectively, and $c$ represents the dimension. After that, we add a linear projection layer to project the features from dimension $c$ to $d$ to match the incremental cross-attention module and reshape the projected features to be patch token $X_P\in \mathbb{R}^{L\times d}$, where $L=hw$. In order to learn new classes while preserve the model performance on the old classes, two learnable tokens are concatenated with the sequence of patch token $X_P$ including a knowledge transfer~(KT) token $x_T\in \mathbb{R}^{d}$ and a knowledge retention (KR) token $x_R \in\mathbb{R}^{d}$. Making the $X_P$, $x_T$ and $x_R$ as the input of the cross-attention module:
\begin{eqnarray}
& & Q = W_q[x_{T}], \nonumber \\
& & K = W_k[x_R,X_P],  \nonumber\\
& &  V = W_v[x_R,X_P], \nonumber\\
& &  z = W_o\mathrm{softmax}\left(\frac{Q K^{T}}{\sqrt{l/h}}\right)V+b_o,
\end{eqnarray}      
where $l$ is the embedding dimension, and $h$ is the number of attention heads. Our incremental cross-attention~(ICA) module defines the KT token~($x_{T}$) as query($Q$). And the concatenation of KR token $x_R$ and patch token $X_P$ (i.e. $[x_R,X_P]$ ) as key($K$) and value($V$). These tokens are fed into the cross-attention~(CA) module:
\begin{eqnarray}
& & e_{1}=x_{T}+\mathrm{CA}(\mathrm{Norm}(x_T,x_R,X_P)), \\
& & e =e_{1}+\mathrm{MLP}(\mathrm{Norm}(e_{1})),
\end{eqnarray}      
where CA($\cdot$) and Norm($\cdot$) denote the cross-attention and layer normalization in \cite{gdtrs}, respectively, and MLP is a multi-layer perception with a single hidden layer. The output embedding $e\in \mathbb{R}^d$ keeps the same dimension as $Q$ (i.e. $x_T$). Then we feed the embedding $e$ into a classification head $\varphi$ and obtain the output logits $o\in \mathbb{R}^{N}$.

\vspace{-0.1cm}
\subsubsection{ICA Based MLCIL}
\vspace{-0.1cm}

In the first session, we add the unified KT token $x_{T}$ which is continuously trained across all the sessions and the KR token $x_{R}^1$ which is only trained on current session. At the session $t$, we expand our ICA module by creating a new KR token $x_R^t$ while keeping the old KR tokens $x_R^{1\sim t-1}$. Therefore, we have one unified KT $x_{T}$ and $t$ KR tokens $x_R^{1\sim t}$ (The old KR tokens $x_R^{1\sim t-1}$ preserve the knowledge of old classes of the corresponding sessions and are frozen at the current session $t$). For each input image, we feed it into the feature extractor $f(;\theta^t)$ and linear projection layer to acquire the patch token $X_P$. By incorporating the KT token with an old session KR token, the ICA outputs a session-specific embedding for $X_P$, which transfers the knowledge of old session to the current session. In order to acquire all the potential object categories of the image, the ICA outputs a total number of t session-specific embeddings $\{e^1,...,e^t\}$ using the current and $t-1$ previous KR tokens, and leverage these embeddings for current session multi-label classification.

Finally, each embedding $e^{1\sim t}$ is fed to the corresponding classification heads $\varphi^{1\sim t}$ with parameters $\phi^{1\sim t}$ to obtain the output logits $o^{1\sim t}$:
\begin{gather}
o^1 =\varphi^1(\mathrm{ICA}((x_{T},x^1_R,X_P);\phi^1)), \nonumber \\
o^2 =\varphi^2(\mathrm{ICA}((x_{T},x^2_R,X_P);\phi^2)), \nonumber \\
\cdots \nonumber  \\
o^t = \varphi^t(\mathrm{ICA}((x_{T},x^t_R,X_P);\phi^t)) ,
\end{gather}
where $o^t \in R^{N^t}$. Then we concatenate all output logits as $O^t=[o^1,o^2,...,o^t]$ to compute the classification loss $\mathbf{L}_{ASL}$. On this basis, to balance stability and plasticity, we concatenate these session-specific embeddings as the output of the ICA module, denoted as $E^t=[e^1,...,e^t]$, to compute the token loss $\mathbf{L}_{TL}$~(see in \ref{loss}). 

\vspace{-0.1cm}
\subsection{Loss Function}
\vspace{-0.1cm}
\label{loss}
Our model is trained on two losses: (1) the classification loss $\mathbf{L}_{ASL}$: asymmetric loss~\cite{asl2020}, and (2) the token loss $\mathbf{L}_{TL}$ applied on the ICA module. In summary, the total loss in the incremental learning (IL) sessions is:
\begin{equation}\label{eq:total}
\mathbf{L}_{IL} = \mathbf{L}_{ASL}  + \lambda \mathbf{L}_{TL},
\end{equation}
where $\lambda$ is hyper-parameter.\\
\textbf{Asymmetric loss:} We adopt an asymmetric loss~\cite{asl2020} for classification. We can predict category probabilities of each image $\mathbf{p} = [p_1, ..., p_{N}]\in \mathbb{R}^{N}$:
\begin{equation}
\label{eq6}
 L_{ASL}=\frac{1}{N} \sum_{n=1}^{N} \left\{
\begin{aligned}
 & (1-p_n)^{\gamma+}log(p_n), & y_n=1, \\
 & p_n^{\gamma-}log(1-p_n), & y_n=0,
\end{aligned}
\right.
\end{equation}
where $\mathbf{y}_n$ is the binary label to indicate if image has label $n$. $\gamma+$ and $\gamma-$ are the positive and negative focusing parameters, respectively. \\
\textbf{Token loss:} To optimize the ICA module and prevent the forgetting of old knowledge, we propose a token loss to penalize the changes of old session-specific embeddings. The $\mathbf{L}_{TL}$ can be written as:
\begin{equation}
  \mathbf{L}_{TL} = 1-<E^{t-1},E^{t}[:e^{t-1}]>,
\end{equation}
where $E^{t-1}$ and $E^{t}$ are the previous and current output of the ICA module, and $<,>$ represents cosine similarity.

\section{Experiment}
\label{sec:experiment}

\begin{figure*}[htb!]
\setlength{\abovecaptionskip}{-0.1cm} 
\setlength{\belowcaptionskip}{-0.3cm} 
\begin{center}
    \includegraphics[width=0.99\textwidth]{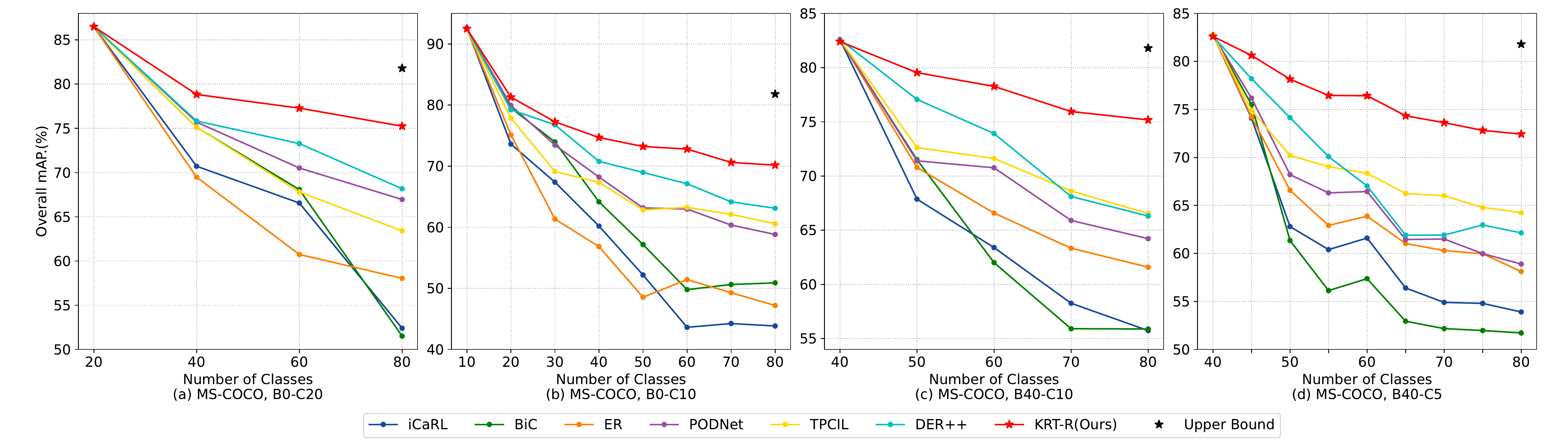}
\end{center}
\caption{Comparison results (mAP\%) on MS-COCO dataset under different protocols against rehearsal-based methods.}
\label{fig:coco}
\end{figure*}

\subsection{Datasets and Experimental Details}

\textbf{Datasets and Benchmark.} We use MS-COCO 2014~\cite{coco2014} and PASCAL VOC 2007~\cite{voc2007} datasets to evaluate the effectiveness of our method in MLCIL task. MS-COCO is a widely-used, large-scale dataset for evaluating multi-label classification. It comprises $122, 218$ images and covers 80 object classes. The training set contains 80K images, the validation set contains 40K images, and on average, each image has 2.9 labels. PASCAL VOC dataset consists of $9,963$ images across 20 object classes with 5K images for training and 5K images for testing. The average number of labels per image is $2.4$. 

Followed by CIL works~\cite{podnet,der+}, we evaluate our methods on MS-COCO dataset with two protocols including 1)\textit{COCO-B0}: we train all 80 classes in several splits including 4 and 8 incremental sessions. 2)\textit{COCO-B40}: we first train a base model on 40 classes and the remaining 40 classes are divided into splits of 4 and 8 sessions. In addition, we evaluate our methods on VOC with two protocols that are 1)\textit{VOC-B0}: this trains the model in batches of 4 classes from scratch. 2)\textit{VOC-B10}: this starts from a model trained on 10 classes, and the remaining 10 classes come in 5 sessions. Inspired by the IOD task~\cite{iod2017,iod2022}, the order of incremental learning is the lexicographical order of category names.

\textbf{Evaluation Metrics.} 
For settings with MLCIL task, we adopt two metrics, average accuracy and last accuracy, which are widely used in CIL works~\cite{podnet,der+}. Following the MLC works~\cite{asl2020,q2l}, we adopt the mean average precision (mAP) to evaluate all the categories that have been learned in each session and report the average mAP (the average of the mAP of all sessions) and the last mAP (final session mAP). To provide a more comprehensive evaluation of the performance after training on all incremental tasks, we also report the per-class F1 measure (CF1) and overall F1-measure (OF1) alongside the last accuracy.

\textbf{Implementation Details.}
Followed by ASL~\cite{asl2020}, we adopt ImageNet-21k pre-trained TResNetM~\cite{tresnet} as our backbone (All compared methods also use ImageNet-21k pre-trained TResNetM or ViT-B/16 as the backbone). We train the model for 20 epochs~(4 warm-up epochs) using Adam~\cite{adam} optimizer and OneCycleLR scheduler with a weight decay of 1e-4. The batch size is set to $64$. To train the base model, we set the learning rate to 4e-5. During the incremental session, we set the learning rate to 1e-4 for COCO and 4e-5 for VOC. The data augmentation techniques include rand augmentation~\cite{augment} as well as cutout~\cite{cutout}. Furthermore, we conduct experiments three times and reported the average results. More implementation details are provided in the \textbf{Appendix}.

\begin{table*}[htb!]
\renewcommand\arraystretch{1.26}
\scriptsize
\setlength\tabcolsep{6.5pt}
\setlength{\abovecaptionskip}{-0.1cm} 
\setlength{\belowcaptionskip}{-0.3cm} 
\begin{center}
\begin{tabular}{l|c|c|cccc|cccc}
\hline
\multirow{3}{*}{\textbf{Method}} & \multirow{2}{*}{\textbf{Source}} & \multirow{3}{*}{\textbf{Buffer size}} & \multicolumn{4}{c|}{\textbf{MS-COCO B0-C10}}  & \multicolumn{4}{c}{\textbf{MS-COCO B40-C10}}    \\ \cline{4-11} 
     &  \multirow{2}{*}{\textbf{Task}}            &  & \multicolumn{1}{c|}{Avg. Acc} & \multicolumn{3}{c|}{Last Acc}& \multicolumn{1}{c|}{Avg. Acc} & \multicolumn{3}{c}{Last Acc} \\ \cline{4-11}  
     &          & & \multicolumn{1}{c|}{mAP (\%)}  &  CF1 & OF1 & mAP (\%)   & \multicolumn{1}{c|}{mAP (\%)} &  CF1 & OF1 & mAP (\%)  \\  \midrule 
    Upper-bound & Baseline &    -     &   \multicolumn{1}{c|}{-}  & 76.4  &  79.4      & 81.8   & \multicolumn{1}{c|}{-}   & 76.4  &  79.4   & 81.8   \\   \midrule \midrule
    FT~\cite{asl2020}& Baseline & \multirow{5}{*}{0}   &  \multicolumn{1}{c|}{38.3} & 6.1 & 13.4 & 16.9 ($\downarrow 49.0$)  & \multicolumn{1}{c|}{35.1} & 6.0 & 13.6 & 17.0 ($\downarrow 57.0$)   \\
    PODNet~\cite{podnet} & CIL &     &  \multicolumn{1}{c|}{43.7} & 7.2 & 14.1 & 25.6 ($\downarrow 40.3$)  & \multicolumn{1}{c|}{44.3} & 6.8 & 13.9 & 24.7 ($\downarrow 49.3$)     \\
     oEWC~\cite{oewc} & CIL& &  \multicolumn{1}{c|}{46.9}    &  6.7    & 13.4  & 24.3 ($\downarrow41.6$) & \multicolumn{1}{c|}{44.8} & 11.1 & 16.5  & 27.3 ($\downarrow 46.7$)\\
     LWF~\cite{LWF}& CIL &     &  \multicolumn{1}{c|}{47.9}    &   9.0   & 15.1  & 28.9 ($\downarrow37.0$) & \multicolumn{1}{c|}{48.6} & 9.5 & 15.8 &  29.9 ($\downarrow 44.1$)                    \\        
     \textbf{KRT}(Ours)& MLCIL&    &  \multicolumn{1}{c|}{\textbf{74.6}} & \textbf{55.6}  & \textbf{56.5} & \textbf{65.9 ($\downarrow 0.0$)} & \multicolumn{1}{c|}{\textbf{77.8}}  & \textbf{64.4}   & \textbf{63.4} & \textbf{74.0 ($\downarrow 0.0$)} \\ 
  \midrule
    TPCIL~\cite{tao2020topology} & CIL &  \multirow{4}{*}{5/class}  &  \multicolumn{1}{c|}{63.8} & 20.1 & 21.6 & 50.8 ($\downarrow 17.5$) &  \multicolumn{1}{c|}{63.1}  &  25.3  & 25.1 & 53.1 ($\downarrow 21.2$)     \\ 
    PODNet~\cite{podnet} & CIL &     &   \multicolumn{1}{c|}{65.7} & 13.6  &17.3 & 53.4 ($\downarrow 14.9$)  & \multicolumn{1}{c|}{65.4} & 24.2   & 23.4 & 57.8 ($\downarrow 16.5 $)   \\
    DER++~\cite{der+}   & CIL &     &  \multicolumn{1}{c|}{68.1} & 33.3 &36.7 & 54.6 ($\downarrow 13.7$) &  \multicolumn{1}{c|}{69.6} & 41.9  & 43.7    & 59.0 ($\downarrow 15.3$)       \\
      \textbf{KRT-R}(Ours) & MLCIL&    & \multicolumn{1}{c|}{\textbf{75.8}}  & \textbf{60.0}  & \textbf{61.0} & \textbf{68.3 ($\downarrow 0.0$)} &  \multicolumn{1}{c|}{\textbf{78.0}} & \textbf{66.0}& \textbf{65.9} &\textbf{74.3 ($\downarrow 0.0$)} \\
  \midrule
    iCaRL~\cite{ICARL} & CIL &   \multirow{7}{*}{20/class}   & \multicolumn{1}{c|}{59.7} & 19.3& 22.8 &  43.8 ($\downarrow 26.4 $)  & \multicolumn{1}{c|}{65.6}  & 22.1& 25.5  & 55.7 ($\downarrow 19.5$)    \\
    BiC~\cite{BIC} & CIL &      & \multicolumn{1}{c|}{65.0} &31.0 &38.1 & 51.1 ($\downarrow 19.1$) & \multicolumn{1}{c|}{65.5}   &  38.1 & 40.7 &  55.9 ($\downarrow 19.3$)      \\
    ER~\cite{ERbase} & CIL &      & \multicolumn{1}{c|}{60.3} & 40.6 & 43.6 & 47.2 ($\downarrow 23.0$)   & \multicolumn{1}{c|}{68.9}  & 58.6 & 61.1 & 61.6 ($\downarrow 13.6$)     \\
    TPCIL~\cite{tao2020topology} & CIL &   &  \multicolumn{1}{c|}{69.4} & 51.7 &52.8 &60.6 ($\downarrow 9.6$) &  \multicolumn{1}{c|}{72.4}  & 60.4   &   62.6     & 66.5 ($\downarrow 8.7$)     \\ 
    PODNet~\cite{podnet} & CIL &    &   \multicolumn{1}{c|}{70.0} & 45.2  &48.7 & 58.8 ($\downarrow 11.4$) & \multicolumn{1}{c|}{71.0} & 46.6   & 42.1 &64.2 ($\downarrow 11.0$)    \\
    DER++~\cite{der+}   & CIL &     &  \multicolumn{1}{c|}{72.7} & 45.2 &48.7 & 63.1 ($\downarrow 7.1$) &  \multicolumn{1}{c|}{73.6} & 51.5   & 53.5       &  66.3 ($\downarrow 8.9$)       \\
      \textbf{KRT-R}(Ours) & MLCIL&    & \multicolumn{1}{c|}{\textbf{76.5}}  & \textbf{63.9}  & \textbf{64.7} & \textbf{70.2 ($\downarrow 0.0$)} &  \multicolumn{1}{c|}{\textbf{78.3}} & \textbf{67.9}& \textbf{68.9} &\textbf{75.2 ($\downarrow 0.0$)} \\
   \midrule
   \midrule
    PRS~\cite{PRS}   & MLOIL & \multirow{3}{*}{1000}     &  \multicolumn{1}{c|}{48.8} & 8.5 & 14.7 & 27.9 ($\downarrow 41.4 $) &  \multicolumn{1}{c|}{50.8} & 9.3    & 15.1   & 33.2 ($\downarrow 41.9 $)         \\
    OCDM~\cite{OCDM} & MLOIL &     &  \multicolumn{1}{c|}{49.5}  & 8.6 & 14.9 & 28.5 ($\downarrow 40.8$)&  \multicolumn{1}{c|}{51.3}  &  9.5   &     15.5      & 34.0 ($\downarrow 41.1$)     \\ 
    \textbf{KRT-R}(Ours) & MLCIL&    & \multicolumn{1}{c|}{\textbf{75.7}}  & \textbf{61.6}  & \textbf{63.6} & \textbf{69.3 ($\downarrow 0.0$)} &  \multicolumn{1}{c|}{\textbf{78.3}} & \textbf{67.5}& \textbf{68.5} &\textbf{75.1 ($\downarrow 0.0$)} \\
   \hline
\end{tabular}
\end{center}
\caption{Class-incremental results on MS-COCO dataset. Compared methods are grouped based on different source tasks. Buffer size 0 means no rehearsal is required, where most SOTA CIL methods are not applicable anymore.}
\label{tb:cocob0c10}
\end{table*}
\vspace{-0.1cm}
\subsection{Comparison Methods}
\vspace{-0.1cm}
For comparative experiments, we run several baselines and state-of-the-art single-label continual learning methods in our MLCIL setting. We select widely recognized and best-performing methods based on several recent CIL works ~\cite{der+,podnet,l2p}. To provide a comprehensive analysis, we also include latest state-of-the-art multi-label online incremental learning methods~\cite{PRS,OCDM}. In addition, we use $\mathbf{L}_{ASL}$ instead of cross-entropy loss as classification loss and rely on the original code base for implementation and hyper-parameter selection to ensure optimal performance.\footnote[2]{Task-incremental learning~\cite{PackNet,hat,ke2020continual,jiang2023neural} and singe-label online incremental learning~\cite{yoon2021online,shim2021online,sun2022information,han2023online,fini2020online} methods are not included as they are not applicable to different class-incremental setting.}


\textbf{Baseline Methods.}
FT method fine-tunes the model without any anti-forgetting constraints. Upper-bound is the supervised training on the data of all tasks, which is usually regarded as the upper-bound performance a IL method can achieve.  

\textbf{Class-incremental Methods.}
We select nine representative CIL works to our MLCIL setting, including oEWC~\cite{oewc}, LWF~\cite{LWF}, iCARL~\cite{ICARL}, BiC~\cite{BIC}, ER~\cite{ERbase}, TPCIL~\cite{tao2020topology}, Der++~\cite{der+}, PODNet~\cite{podnet}, and L2P~\cite{l2p}. oEWC~\cite{oewc} and LwF~\cite{LWF} are representative regularization-based works. ICARL, BiC and ER are classical rehearsal-based methods. Der++, PODNet, and TPCIL are best-performing rehearsal-based methods. L2P~\cite{l2p} is the latest SOTA CIL method based on ViT-B/16, we compare the relative performance to the corresponding upper-bound performance for fairness.

\textbf{ML Online-incremental Methods.}
We select two latest SOTA ML online incremental learning methods in our MLCIL setting to compare, including PRS~\cite{PRS} and OCDM~\cite{OCDM}.

\textbf{Our Methods.} KRT is our proposed method without rehearsal buffer. KRT-R is KRT equipped with a rehearsal buffer for a fair comparison with SOTA methods.
\vspace{-0.1cm}
\subsection{Comparison Results}
\vspace{-0.1cm}
\textbf{Results on MS-COCO.}
Table~\ref{tb:cocob0c10} shows the results on MS-COCO B0-C10 and B40-C10 benchmarks. KRT outperforms all comparing methods consistently, in terms of both average accuracy (mAP) and last accuracy (CF1, OF1 and mAP). Specifically, when the buffer size is large (20/class), our method achieves the best final accuracy of \textbf{70.2\%} and \textbf{75.2\%} on two benchmarks, which outperforms the latest SOTA rehearsal-based methods by \textbf{7.1\%} and \textbf{8.7\%}, respectively. When the buffer size gets smaller (5/class), KRT-R achieves even greater performance gains compared to other continual learning methods. It is worth noting that when the buffer size is set to 0, rehearsal-based CIL methods become ineffective. For example, the final mAP of the PODNet dropped sharply by at least \textbf{33.2\%}. However, KRT still maintains superior performance by outperforming the regularization-based methods and other rehearsal-based methods even they have large rehearsal buffer.
\begin{table}[t!]
\renewcommand\arraystretch{1.26}
\scriptsize
\setlength\tabcolsep{3pt}
\setlength{\abovecaptionskip}{-0.1cm} 
\setlength{\belowcaptionskip}{-0.4cm}
\begin{center}
\begin{tabular}{l|c|c|cc}
\hline
\multirow{2}{*}{\textbf{Method}} &\multirow{2}{*}{\textbf{Backbone}} &\multirow{2}{*}{\textbf{Param.}} & \textbf{Avg.} & \textbf{Last}  \\
                            & &            & \textbf{mAP\%}  & \textbf{mAP\%}      \\ \midrule
    Upper-bound         & \multirow{3}{*}{ViT-B/16}& \multirow{3}{*}{86.0M}   &  -   &      83.16    \\
    L2P~\cite{l2p}      &                      &     &   73.07        &  70.42   ($\nabla$ 12.74  )      \\
    L2P-R~\cite{l2p}    &                  &   &  73.64        &  71.68  ($\nabla$ 11.48 )          \\ \midrule
   Upper-bound     &  \multirow{3}{*}{TResNetM}& \multirow{3}{*}{29.4M}   &  -      & 81.80      \\ 
   \textbf{KRT}(Ours)     &                  &       &\textbf{77.83}   &   \textbf{74.02} ($\nabla$ \textbf{7.78} )   \\
   \textbf{KRT-R}(Ours)      &                 &    &\textbf{78.34}    &    \textbf{75.18} ($\nabla$ \textbf{6.62} )   \\ \hline 
\end{tabular}
\end{center}
\caption{Class-Incremental results on MS-COCO dataset under the B40-C10 setting against prompt-based CIL method. $\nabla$ indicates the gap towards the Upper Bound of corresponding backbone.}
\label{tb:l2p}
\end{table}

Figure~\ref{fig:coco} shows the comparison curves on four challenge benchmarks with larger buffer size. It is observed that KRT-R(Ours) consistently outperforms all other CIL methods at every session regardless of the incremental settings and is the closest to the Upper Bound. As the number of sessions increases, we observe a widening gap between the performance of the KRT method and other methods. This suggests that our method is better suited for long-term incremental learning scenarios. 

Table~\ref{tb:cocob0c10} also presents a comparison between KRT-R and MLOIL methods. We observe that the online learning methods do not perform well on the MLCIL task. Our KRT outperforms both PRS and OCDM by a large margin.

Table~\ref{tb:l2p} shows the comparison between KRT and prompt-based methods. Since the L2P method is based on the pre-trained ViT, we use the towards to the upper bound ($\nabla$) to measure the performance of each method given a specific backbone. We can observe that KRT relatively outperforms L2P by at least \textbf{4.86\%} with or without rehearsal buffer.

\begin{table}[t!]
\renewcommand\arraystretch{1.26}
\scriptsize
\setlength{\abovecaptionskip}{-0.1cm} 
\setlength{\belowcaptionskip}{-0.4cm} 
\setlength\tabcolsep{3pt}
\begin{center}
\begin{tabular}{l|c|cc|cc} \hline
\multirow{2}{*}{\textbf{Method}} & \textbf{Buffer} & \multicolumn{2}{c|}{\textbf{VOC B0-C4}} & \multicolumn{2}{c}{\textbf{VOC B10-C2}} \\ \cline{3-6} 
         &        \textbf{Size}          &     Avg. Acc    &   Last Acc       &    Avg. Acc        &  Last Acc           \\ \midrule
      Upper bound     &  \multirow{2}{*}{-}          &       -    &  93.6        &    -       &     93.6     \\
       FT~\cite{asl2020}  &   & 82.1& 62.9  & 70.1  & 43.0       \\  \midrule
       iCarL~\cite{ICARL} & \multirow{7}{*}{2/class} &  87.2 & 72.4 ($\downarrow11.0$)  & 79.0 & 66.7 ($\downarrow13.8$) \\
        BIC~\cite{BIC}  &       & 86.8 & 72.2 ($\downarrow11.2 $)   & 81.7  &69.7 ($\downarrow10.8$)   \\  
        ER~\cite{ERbase}  &    & 86.1   & 71.5 ($\downarrow11.9$)   & 81.5 & 68.6 ($\downarrow11.9$)  \\  
         TPCIL~\cite{tao2020topology}  &  & 87.6  & 77.3 ($\downarrow6.1$)   & 80.7   &70.8 ($\downarrow9.7$)  \\  
        PODNet~\cite{podnet}  &   &  88.1  & 76.6 ($\downarrow6.8$)   &  81.2   & 71.4 ($\downarrow9.1$) \\  
        DER++~\cite{der+}  &     &  87.9  & 76.1 ($\downarrow7.3$)  & 82.3   &70.6 ($\downarrow9.9$)  \\  
        \textbf{KRT-R}(Ours)  &   &   \textbf{90.7}  & \textbf{83.4} ($\downarrow0.0$)  & \textbf{87.7} & \textbf{80.5}($\downarrow0.0$)
 \\  \hline
\end{tabular}
\end{center}
\caption{Comparison results on PASCAL VOC dataset. All metric are in mAP\%}
\label{tb:voc}
\end{table}
\textbf{Results on PASCAL VOC.}
Table~\ref{tb:voc} summarizes the experimental results on PASCAL VOC dataset. We observe a similar conclusion to those on MS-COCO dataset. Concretely, KRT consistently surpasses other methods by a considerable margin on two benchmarks. In the comparison results with the incremental data split into 5 sessions, KRT achieves the best last mAP value of \textbf{83.4}\% and outperforms the other methods by \textbf{6.1\%(77.3\%$\xrightarrow{}$83.4\%)}. Moreover, on the B10-C2 benchmark, our method outperforms second best method from \textbf{71.4\%} to \textbf{80.5\%}(\textbf{9.1\%}) at the last session.

The outstanding performance of KRT over all compared methods on two MLC datasets indicates that the effectiveness of our methods for improving recognition performance and mitigating forgetting for MLCIL task even without a rehearsal buffer.

\vspace{-0.1cm}
\subsection{Ablation Study}
\label{ablation}
\vspace{-0.1cm}
\textbf{The Effectiveness of Each Component.} 
 Table~\ref{tb:abalation} demonstrates the results of our ablative experiments on COCO B40-C10 setting with large buffer size. We use a distillation loss $\mathbf{L}_{KD}$~\cite{lucir} applied on the globally pooled feature as the \textbf{baseline} method and generate three additional variants of KRT. (a) KRT w/o DPL: We optimize using only the ICA module. (b) KRT w/o ICA: We optimize using only the DPL module. (c) KRT w/ KD: we add extra the loss $\mathbf{L}_{KD}$ to our KRT method.
 \begin{table}[htb]
\renewcommand\arraystretch{1.35}
\scriptsize
\setlength{\abovecaptionskip}{-0.1cm} 
\setlength{\belowcaptionskip}{-0.2cm} 
\begin{center}
\setlength\tabcolsep{6pt}
\begin{tabular}{lcccccccccc}
\hline
Model & KD & ICA & DPL & Avg. Acc & Last Acc   \\
\hline
Baseline & \checkmark & & & 65.93  & 58.02 ($\uparrow 0.0 $)  \\\hline
(a) w/o DPL &   & \checkmark  &  &  77.06 & 71.97 ($\uparrow 13.95$)  \\
(b) w/o ICA &   &   & \checkmark  &  77.14  & 73.12 ($\uparrow 15.10$)   \\
(c) w/ KD & \checkmark & \checkmark  & \checkmark&  78.14 & 74.77 ($\uparrow 16.75$)   \\ \hline
 \textbf{KRT}  &  & \checkmark & \checkmark &  \textbf{78.34} & \textbf{75.18 ($\uparrow 17.16$)  }\\
\hline
\end{tabular}
\end{center}
\caption{The contribution of each component.}
\label{tb:abalation}
\end{table}

As shown in Table~\ref{tb:abalation}, the baseline model produces the lowest last mAP of \textbf{58.02\%}. Using ICA or DPL modules separately both bring a significant improvement (rows(a,b)). Only using ICA module (row a) improves the last mAP by \textbf{13.95\%} and with DPL module used separately (row b), we observe a \textbf{15.10\%} relative improvement. Applying the KD loss degrades the performance (row c). Though it is popularly used by CIL methods~\cite{lucir,podnet}, it may be not so effective for MLCIL. These results strongly prove that the ICA and DPL module are very effective to prevent forgetting and improve performance for the MLCIL tasks.

\begin{figure}[t]
\setlength{\abovecaptionskip}{-0.1cm} 
\setlength{\belowcaptionskip}{-0.2cm} 
\small
\begin{center}
    \subfloat[]{
        \label{fig:hp}
        \includegraphics[width=0.23\textwidth]{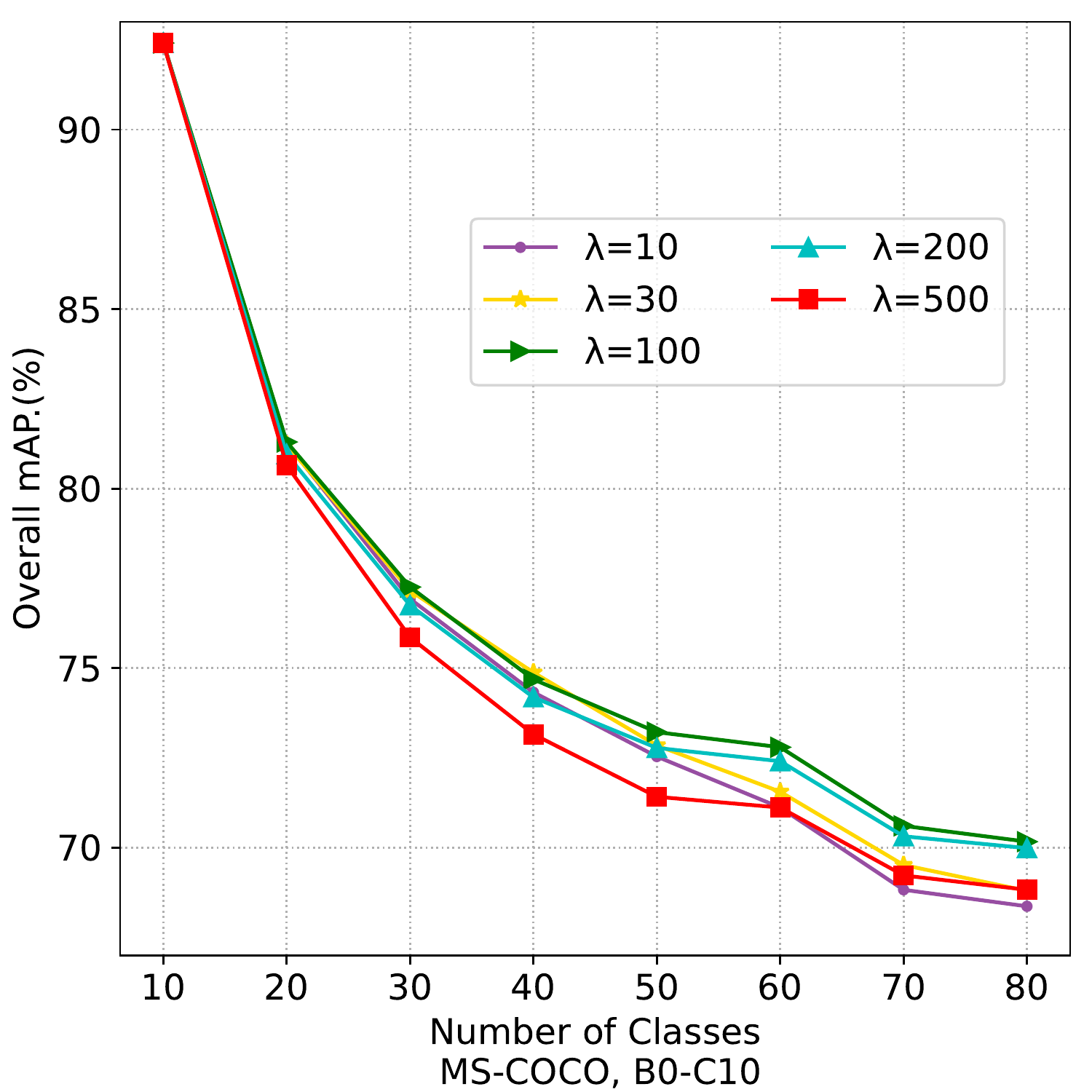}}
    \subfloat[]{
        \label{fig:hp1}
        \includegraphics[width=0.23\textwidth]{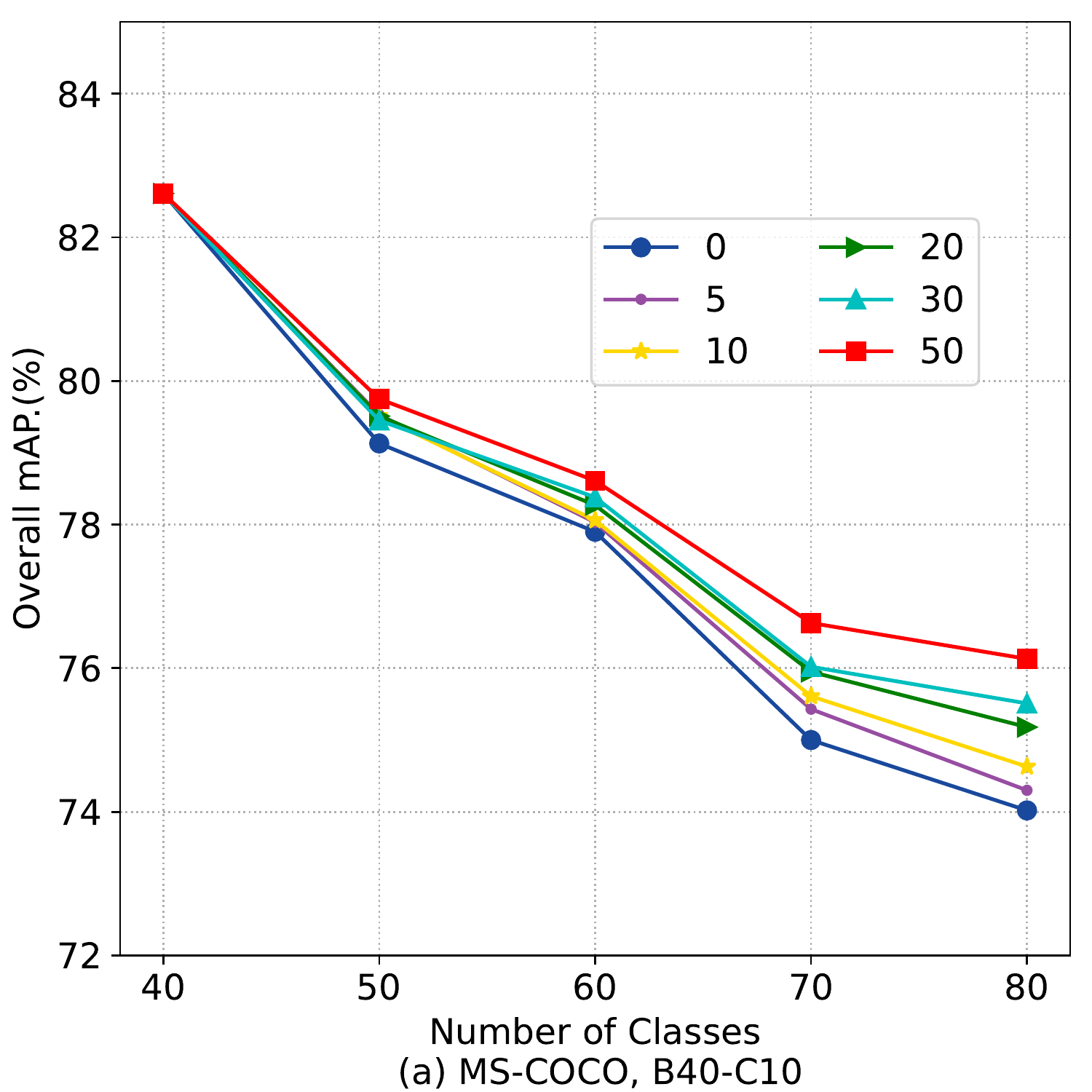}}
\end{center}
\caption{\textbf{Robustness Testing.} (a) Sensitive study of hyper-parameter $\lambda$. (b) The influence of buffer size.}
\label{fig:manifold}
\end{figure}

\textbf{Sensitive Study of Hyper-parameter $\lambda$.} To verify the robustness of KRT , we conduct experiments on MS-C0C0 B0-C10 with different hyper-parameters $\lambda$. More specifically, we test~$\lambda=10,30,100,200,500$ respectively. The comparison results are shown in Figure~\ref{fig:hp}. We can see that our KRT get best performance when $\lambda=100$ and the performance changes are minimal under different $\lambda$. 

\textbf{The Influence of Buffer Size.} We gradually increase the buffer size from 0 per class to 50 per class and report the performance of the our KRT on MS-COCO B40-C10 in Figure~\ref{fig:hp1}. The final mAP only increases from \textbf{74.02\%} to \textbf{76.13\%} as the buffer size change from 0 to 50. Form the results. we can see that our KRT is more effective and robust and it can overcome forgetting even without buffer. 

\begin{figure}[t]
\setlength{\abovecaptionskip}{-0.15cm} 
\setlength{\belowcaptionskip}{-0.15cm} 
\begin{center}
    \includegraphics[width=0.45\textwidth]{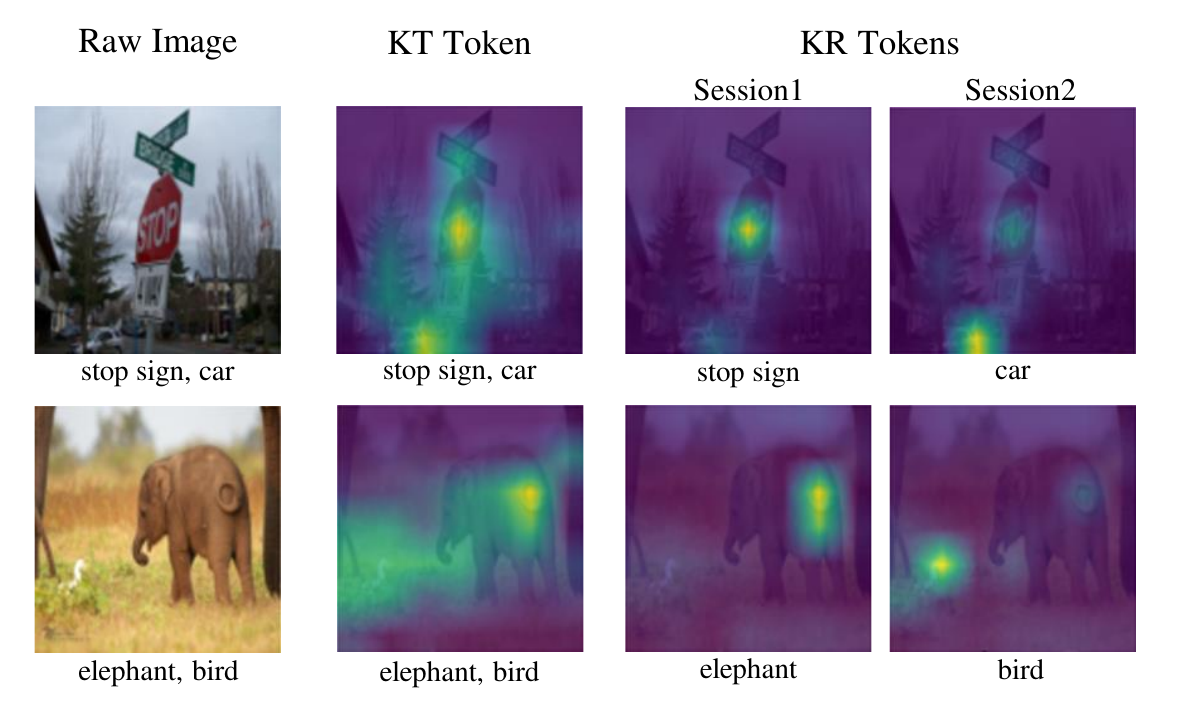}
\end{center}
\caption{Visualization of ICA module.}
\label{fig:vis}
\end{figure}

\textbf{Visualization of ICA Module.}
To further demonstrate the effectiveness of ICA module, we illustrate several attention map examples of the KR and KT tokens in Figure~\ref{fig:vis}. The KR tokens of different sessions only maintains the category knowledge of the current session, and the continuously trained KT token learns the knowledge of all sessions.

More detailed experimental results and more visualization images are provided in the \textbf{Appendix}.

\vspace{-0.1cm}
\section{Conclusion}
\vspace{-0.1cm}
\label{sec:conclusion}
In this paper, we focus on a challenging but more practical problem named multi-label class incremental learning (MLCIL). Compared to the vanilla CIL problem, MLCIL are faced with two major challenges: the \emph{label absence} of old classes and the \emph{information dilution} during knowledge transfer. To solve these challenges, we propose a knowledge restore and transfer (KRT) framework containing two key components, \ie, a {dynamic pseudo-label} (DPL) module to solve the label absence problem by restoring the knowledge of old classes to the new data and an incremental cross-attention~(ICA) module with session-specific KR tokens storing knowledge and a unified KT token transferring knowledge to solve the information dilution problem. Extensive experimental results on MS-COCO and PASCAL VOC datasets show that our method significantly outperforms existing state-of-the-art methods and demonstrate the superiority of the proposed method.

\vspace{-0.1cm}
\section*{Acknowledgments}
\vspace{-0.1cm}
This work was funded by the National Key Research and Development Project of China under Grant No. 2020AAA0105600, and by the National Natural Science Foundation of China under Grant No. U21B2048 and No. 62006183. Thanks to Huawei's support.

\appendix
\label{sec:appendix}
\section{Appendix}

\subsection{Other Related Work}

\textbf{Incremental object detection}(IOD) applies incremental learning to object detection specifically. Both KD and ER have been applied to IOD task and implement on different detectors. \cite{iod2017} first uses the KD to the output of Faster R-CNN and subsequent methods~\cite{iod2022,IOD2021} add KD terms on the intermediate feature maps and region proposal networks or store a set of exemplars to fine-tune the model. In addition to being applied to CNN detectors, ER and KD have also been applied to the transformer network DETR~\cite{IOD2023}.

\textbf{Incremental Semantic Segmentation}(ISS) methods can be classified into regularization-based and replay-based approaches. The former approaches such as SDR~\cite{SDR} and PLOP~\cite{PLOP} propose different KD strategies to regularize a current model in a latent feature space. The second approaches~\cite{iss-2021er,iss-2021er2} rely on an ER strategy, involving retention of a small set of exemplars or pseudo information for previous categories.

It is evident that the IOD and ISS methods are not directly applicable to MLCIL tasks due to their reliance on specific detection and segmentation frameworks. Therefore, conducting research on MLCIL with only image-level annotations is of great value and significance.

\subsection{More Experimental Details}

All models are implemented with PyTorch and trained on 2 RTX 3090 GPUS. We resize images to $h\times w=224 \times 224$ as the input resolution and the size of output feature is $7\times7\times2048$. The extracted features are fed into the ICA module after linear projection and adding position encodings. We set the dimension $d=384$ for COCO and $d=768$ for VOC datasets. For the ICA module, the embedding dimension $l$ is set to $384$ for COCO and $768$ for VOC datasets, and the number of heads $h$ is set to $8$. For the DPL method, the threshold $\eta$ is initialized as $0.8$, and the target value $\mu$ is set to $2.9$ for COCO and $1.4$ for VOC datasets. The sensitive study of hyper-parameter $\lambda$ in ablation study has summarized that the performance of our methods changes are minimal under different $\lambda$ and we report the best hyper-parameter values under different protocols. Concretely, the $\lambda$ is set to 100 for B0 benchmark and 300 for B40(B10) benchmark. 

All compared methods, including baselines, CIL methods~\cite{LWF,oewc,ERbase,ICARL,EEIL,der+,podnet,tao2020topology}, and MLOIL methods~\cite{OCDM,PRS}, utilize TResNetM pre-trained on ImageNet-21k as the backbone (L2P~\cite{l2p} employs ViT-B/16 pre-trained on ImageNet-21k as the backbone). To adapt SCIL methods for MLCIL tasks, we employ $\mathbf{L}_{ASL}$ as the classification loss instead of cross-entropy loss and rely on the original codebase to implement the method and carefully select hyper-parameters to ensure optimal performance. For the MLOIL methods~\cite{OCDM,PRS}, we directly implement them in MLCIL protocol using their original codebase.
\begin{algorithm}[htbp]
\small
\setstretch{1.35}
\caption{Dynamic Pseudo-Label}
\label{Algorithm1}
\begin{algorithmic}[1]
\REQUIRE Session $t$ training set $\mathbf{X}^{t}$, Initial threshold $\eta$;
\REQUIRE Session $t$ target value $\mu^t$, Old model $\Theta^{t-1}$.
\ENSURE Updated training set $\hat{\mathbf{X}}^{t}$. 
\STATE Employ model $\Theta^{t-1}$ to infer the training set $\mathbf{X}^{t}$ based on initial threshold $\eta$ to obtain pseudo-label set $\mathbf{S}^t$ 
\STATE Count the number of images ${M^t}$ in training set $\mathbf{X}^{t}$ 
\STATE Calculate the the average pseudo labels per image $\beta^t=\frac{|\mathbf{S}^t|}{M^t}$ 
\WHILE{(\,\,$|\beta^t-\mu^t|\,>\,$$1e^{-1}$)\.} 
\IF{$\beta^t >\mu^t$} 
		\STATE $\eta^t=\eta^t+1e-2$
		\ELSE[$\beta^t \leq \mu^t$] \STATE $\eta^t=\eta^t-1e-2$
		\ENDIF 
\STATE Employ model $\Theta^{t-1}$ to infer the training set $\mathbf{X}^{t}$ based on $\eta^t$ to obtain pseudo-label set $\mathbf{S}^t$ 
\STATE Calculate the the average pseudo labels per image $\beta^t=\frac{|\mathbf{S}^t|}{M^t}$ 
\ENDWHILE
\STATE Merge the pseudo-label set $\mathbf{S}^t$ with the label set $\mathbf{Y}^t$ as new ground truth $\hat{\mathbf{Y}}^t$ and obtain the updated training set $\hat{\mathbf{X}}^{t}$
\end{algorithmic}
\end{algorithm}

\subsection{Discussion of ICA Parameters}

The initial parameter count of the ICA module is 2.4M including the Linear Projection and MHSA and MLP components. In the MLCIL task, we add a KT token at the first session and a KR token at each session. Assuming there are a total number of 8 sessions, the amount of extra parameters increased by only 9*384=3,456 (0.003M). Compared with the overall parameters (about 30M), extra adding parameters almost could be ignored. Therefore, there is no issue of excessive additional parameters caused by the increment of sessions, instead, our method is suited for long-term incremental learning scenarios.

\subsection{The Algorithms of DPL Module}

  As mentioned in our main paper, the algorithm of DPL is presented in Algorithm~\ref{Algorithm1}.

\subsection{More Comparison Results and Detailed }

  In order to demonstrate that KRT can achieve MLCIL tasks without any pre-trained, effectively learning new classes, we train KRT from scratch on a completely No pre-trained model and compare with FT approach. The results in Table \ref{tab:nopre} show that KRT exhibits significant improvements even \textbf{without any pre-trained} in the model.
   
\begin{table}[htbp]
\centering
\scriptsize
\renewcommand\arraystretch{1.3}
\renewcommand\tabcolsep{4pt}
\footnotesize
\centering
\scalebox{0.9}{
\begin{tabular}{c|cc|cc|c}
\hline
\multicolumn{1}{c|}{\textbf{Pre-trained}} & \multicolumn{2}{c|}{FT (Baseline)} & \multicolumn{2}{c|}{KRT (Ours)} & \multirow{2}{*}{Upper-bound } \\
 \multicolumn{1}{c|}{\textbf{Model}}&\multicolumn{1}{c}{\textbf{Last Acc}} & \multicolumn{1}{c|}{\textbf{Avg Acc}} &\multicolumn{1}{c}{\textbf{Last Acc}} & \multicolumn{1}{c|}{\textbf{Avg Acc}} &  \\ \hline
 No & 9.60 & 21.37 & \textbf{48.38} & \textbf{52.09} & 62.29 \\
\hline
\end{tabular}}
\caption{Results (mAP\%) on MS-COCO under the B40-C10.}
\label{tab:nopre}
\end{table}

\begin{figure}[t]
\setlength{\abovecaptionskip}{-0.2cm} 
\begin{center}
    \includegraphics[width=0.5\textwidth]{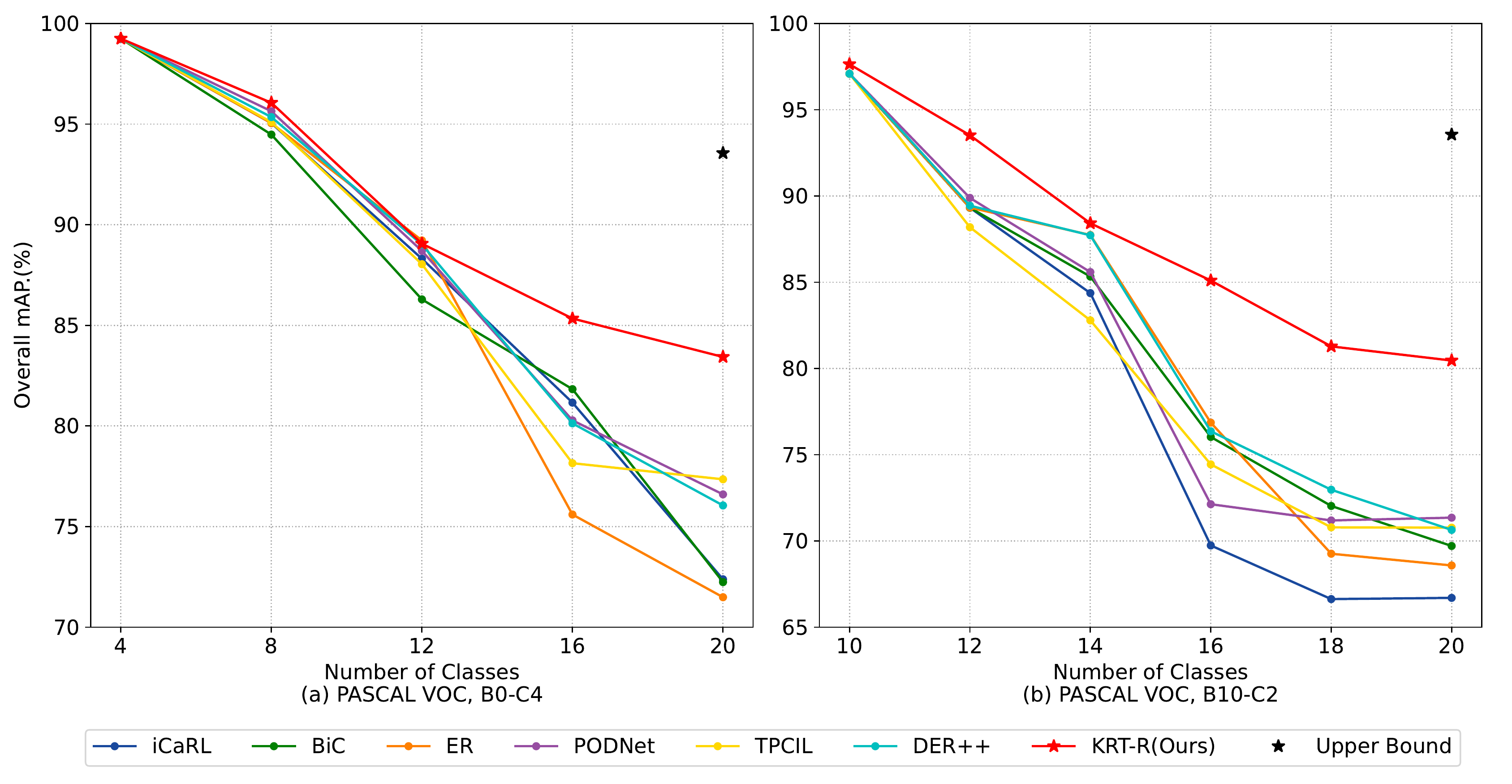}
\end{center}
\caption{Comparison results (mAP\%) on PASCAL VOC.}
\label{fig:voc}
\end{figure}

Moreover, Figure~\ref{fig:voc} presents a comparison of VOC curves on B0C4 and B10C2 benchmarks. Table~\ref{tb:b0c20} and Table~\ref{tb:b0c10} provide detailed per-session performance of different methods on COCO B0C20 and B0C10 benchmarks, respectively. Similarly, Table~\ref{tb:b40c10} and Table~\ref{tb:b40c5} illustrate the per-session performance of various methods on COCO B40C10 and B40C5 benchmarks. Additionally, Table~\ref{tb:b0c4} and Table~\ref{tb:b10c2} display detailed per-session performance of different methods on the VOC B0C4 and B10C2 benchmarks.

\subsection{More Visualization Results}

\begin{figure*}[htb!]
\begin{center}
    \includegraphics[width=0.8\textwidth]{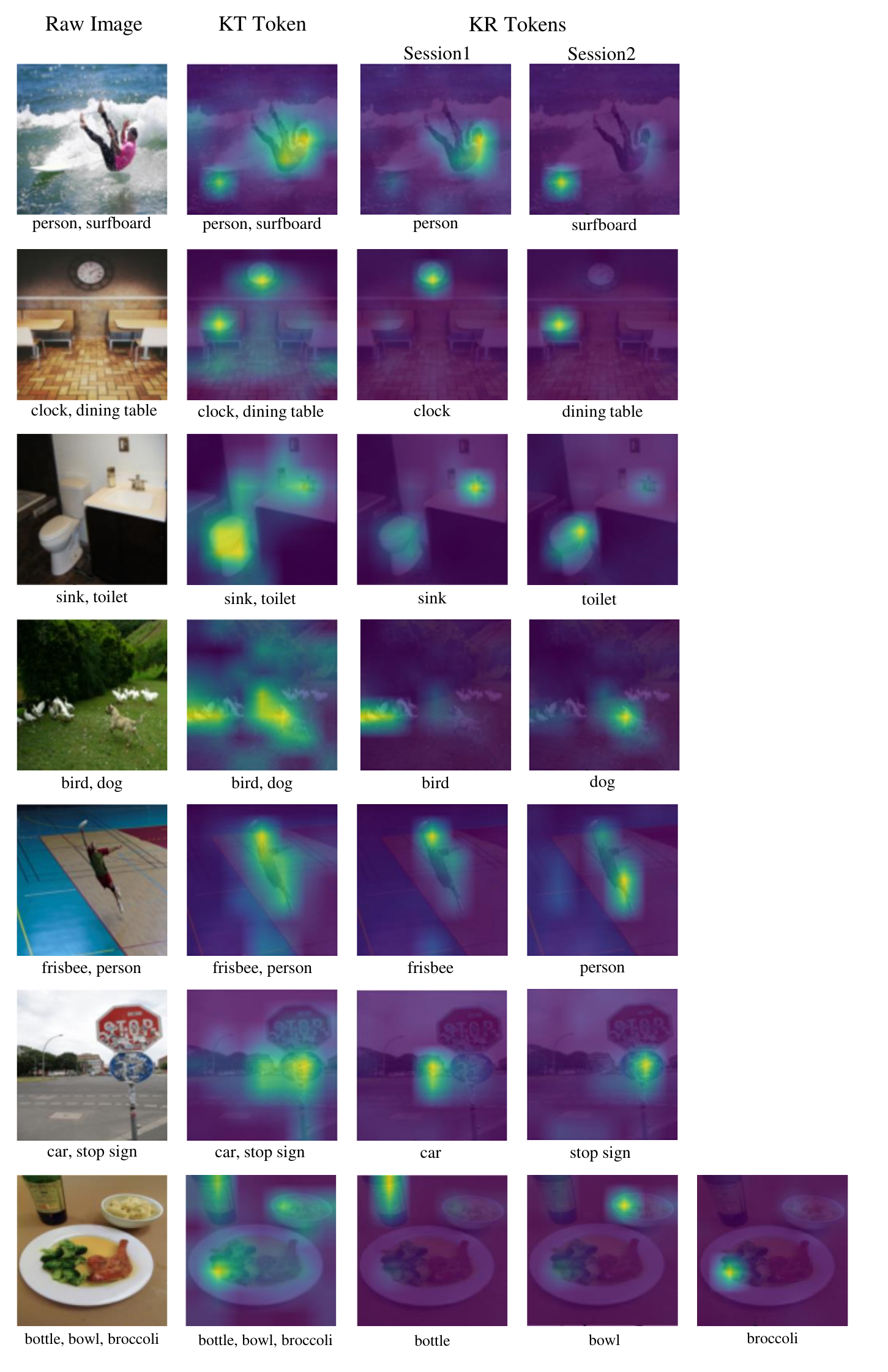}
\end{center}
\caption{Visualization of ICA module.}
\label{fig:app1}
\end{figure*}

We provide more visualization results of cross-attention maps examples of the KR and KT tokens in Figure~\ref{fig:app1}

{\small
\bibliographystyle{ieee_fullname}
\bibliography{egbib}

\begin{thebibliography}{10}\itemsep=-1pt

\bibitem{CCGN}
Davide Abati, Jakub Tomczak, Tijmen Blankevoort, Simone Calderara, Rita
  Cucchiara, and Babak~Ehteshami Bejnordi.
\newblock Conditional channel gated networks for task-aware continual learning.
\newblock In {\em CVPR}, pages 3931--3940, 2020.

\bibitem{MAS}
Rahaf Aljundi, Francesca Babiloni, Mohamed Elhoseiny, Marcus Rohrbach, and et
  al.
\newblock Memory aware synapses: Learning what (not) to forget.
\newblock In {\em ECCV}, pages 139--154, 2018.

\bibitem{ashok2022class}
Arjun Ashok, KJ Joseph, and et al.
\newblock Class-incremental learning with cross-space clustering and controlled
  transfer.
\newblock In {\em ECCV}, pages 105--122. Springer, 2022.

\bibitem{il2m}
Eden Belouadah and et al.
\newblock Il2m: Class incremental learning with dual memory.
\newblock In {\em ICCV}, pages 583--592, 2019.

\bibitem{der+}
Pietro Buzzega, Matteo Boschini, Angelo Porrello, and et al.
\newblock Dark experience for general continual learning: a strong, simple
  baseline.
\newblock {\em NIPS}, 33:15920--15930, 2020.

\bibitem{EEIL}
Francisco~M Castro, Manuel~J Mar{\'\i}n-Jim{\'e}nez, Nicol{\'a}s Guil, Cordelia
  Schmid, and Karteek Alahari.
\newblock End-to-end incremental learning.
\newblock In {\em ECCV}, pages 233--248, 2018.

\bibitem{co2l}
Hyuntak Cha, Jaeho Lee, and Jinwoo Shin.
\newblock Co2l: Contrastive continual learning.
\newblock In {\em ICCV}, pages 9516--9525, 2021.

\bibitem{iss-2021er}
Sungmin Cha, YoungJoon Yoo, and et al.
\newblock Ssul: Semantic segmentation with unknown label for exemplar-based
  class-incremental learning.
\newblock {\em NIPS}, 34:10919--10930, 2021.

\bibitem{chen2021learning}
Zhaomin Chen, Xiu-Shen Wei, Peng Wang, and Yanwen Guo.
\newblock Learning graph convolutional networks for multi-label recognition and
  applications.
\newblock {\em In TPAMI}, 2021.

\bibitem{augment}
Ekin~D Cubuk, Barret Zoph, Jonathon Shlens, and Quoc~V Le.
\newblock Randaugment: Practical automated data augmentation with a reduced
  search space.
\newblock In {\em CVPR Workshops}, pages 702--703, 2020.

\bibitem{cutout}
Terrance DeVries and Graham~W Taylor.
\newblock Improved regularization of convolutional neural networks with cutout.
\newblock {\em arXiv preprint arXiv:1708.04552}, 2017.

\bibitem{lwm}
Prithviraj Dhar, Rajat~Vikram Singh, Kuan-Chuan Peng, Ziyan Wu, and Rama
  Chellappa.
\newblock Learning without memorizing.
\newblock In {\em CVPR}, pages 5138--5146, 2019.

\bibitem{dong2021few}
Songlin Dong, Xiaopeng Hong, Xiaoyu Tao, Xinyuan Chang, Xing Wei, and Yihong
  Gong.
\newblock Few-shot class-incremental learning via relation knowledge
  distillation.
\newblock In {\em AAAI}, volume~35, pages 1255--1263, 2021.

\bibitem{PLOP}
Arthur Douillard, Yifu Chen, and et al.
\newblock Plop: Learning without forgetting for continual semantic
  segmentation.
\newblock In {\em CVPR}, Jun 2021.

\bibitem{podnet}
Arthur Douillard, Matthieu Cord, and et al.
\newblock Podnet: Pooled outputs distillation for small-tasks incremental
  learning.
\newblock In {\em ECCV}, pages 86--102. Springer, 2020.

\bibitem{dytox}
Arthur Douillard, Alexandre Ram{\'e}, Guillaume Couairon, and Matthieu Cord.
\newblock Dytox: Transformers for continual learning with dynamic token
  expansion.
\newblock In {\em CVPR}, pages 9285--9295, 2022.

\bibitem{AGCN}
Kaile Du, Fan Lyu, Fuyuan Hu, Linyan Li, Wei Feng, Fenglei Xu, and Qiming Fu.
\newblock Agcn: augmented graph convolutional network for lifelong multi-label
  image recognition.
\newblock In {\em ICME}, pages 01--06. IEEE, 2022.

\bibitem{voc2007}
Mark Everingham, Luc Van~Gool, Christopher~KI Williams, John Winn, and Andrew
  Zisserman.
\newblock The pascal visual object classes (voc) challenge.
\newblock {\em International Journal of Computer Vision}, 88(2):303--338, 2010.

\bibitem{iod2022}
Tao Feng, Mang Wang, and et al.
\newblock Overcoming catastrophic forgetting in incremental object detection
  via elastic response distillation.
\newblock In {\em CVPR}, pages 9427--9436, 2022.

\bibitem{fini2020online}
Enrico Fini, St{\'e}phane Lathuiliere, Enver Sangineto, Moin Nabi, and Elisa
  Ricci.
\newblock Online continual learning under extreme memory constraints.
\newblock In {\em ECCV}, pages 720--735. Springer, 2020.

\bibitem{MCRN2013}
Yunchao Gong, Yangqing Jia, Thomas Leung, Alexander Toshev, and Sergey Ioffe.
\newblock Deep convolutional ranking for multilabel image annotation.
\newblock {\em arXiv preprint arXiv:1312.4894}, 2013.

\bibitem{guo2020online}
Qiushan Guo, Xinjiang Wang, Yichao Wu, Zhipeng Yu, Ding Liang, Xiaolin Hu, and
  Ping Luo.
\newblock Online knowledge distillation via collaborative learning.
\newblock In {\em CVPR}, pages 11020--11029, 2020.

\bibitem{han2023online}
Ya-nan Han and Jian-wei Liu.
\newblock Online continual learning via the knowledge invariant and spread-out
  properties.
\newblock {\em Expert Systems with Applications}, 213:119004, 2023.

\bibitem{lucir}
Saihui Hou, Xinyu Pan, Chen~Change Loy, Zilei Wang, and Dahua Lin.
\newblock Learning a unified classifier incrementally via rebalancing.
\newblock In {\em CVPR}, pages 831--839, 2019.

\bibitem{jiang2023neural}
Jian Jiang and Oya Celiktutan.
\newblock Neural weight search for scalable task incremental learning.
\newblock In {\em Proceedings of the IEEE/CVF Winter Conference on Applications
  of Computer Vision}, pages 1390--1399, 2023.

\bibitem{IOD2021}
K~J Joseph, Salman Khan, Fahad~Shahbaz Khan, and Vineeth~N Balasubramanian.
\newblock Towards open world object detection.
\newblock In {\em CVPR}, Jun 2021.

\bibitem{kang2022class}
Minsoo Kang, Jaeyoo Park, and Bohyung Han.
\newblock Class-incremental learning by knowledge distillation with adaptive
  feature consolidation.
\newblock In {\em CVPR}, pages 16071--16080, 2022.

\bibitem{ke2020continual}
Zixuan Ke, Bing Liu, and Xingchang Huang.
\newblock Continual learning of a mixed sequence of similar and dissimilar
  tasks.
\newblock {\em NIPS}, 33:18493--18504, 2020.

\bibitem{PRS}
Chris~Dongjoo Kim and et al.
\newblock Imbalanced continual learning with partitioning reservoir sampling.
\newblock In {\em ECCV}, 2020.

\bibitem{adam}
Diederik~P Kingma and Jimmy Ba.
\newblock Adam: A method for stochastic optimization.
\newblock {\em arXiv preprint arXiv:1412.6980}, 2014.

\bibitem{EWC}
James Kirkpatrick, Razvan Pascanu, Neil Rabinowitz, Joel Veness, Guillaume
  Desjardins, Andrei~A Rusu, Kieran Milan, John Quan, Tiago Ramalho, Agnieszka
  Grabska-Barwinska, et~al.
\newblock Overcoming catastrophic forgetting in neural networks.
\newblock {\em National Academy of Sciences}, 114(13):3521--3526, 2017.

\bibitem{C-Trans}
Jack Lanchantin, Tianlu Wang, Vicente Ordonez, and Yanjun Qi.
\newblock General multi-label image classification with transformers.
\newblock In {\em CVPR}, pages 16478--16488, 2021.

\bibitem{LWF}
Zhizhong Li and Derek Hoiem.
\newblock Learning without forgetting.
\newblock {\em IEEE Transactions on Pattern Analysis and Machine Intelligence},
  40(12):2935--2947, 2017.

\bibitem{OCDM}
Yan-Shuo Liang and Wu-Jun Li.
\newblock Optimizing class distribution in memory for multi-label online
  continual learning.
\newblock {\em arXiv preprint arXiv:2209.11469}, 2022.

\bibitem{focal2017}
Tsung-Yi Lin, Priya Goyal, Ross Girshick, Kaiming He, and Piotr Doll{\'a}r.
\newblock Focal loss for dense object detection.
\newblock In {\em ICCV}, pages 2980--2988, 2017.

\bibitem{coco2014}
Tsung-Yi Lin, Michael Maire, Serge Belongie, James Hays, Pietro Perona, Deva
  Ramanan, Piotr Doll{\'a}r, and C~Lawrence Zitnick.
\newblock Microsoft coco: Common objects in context.
\newblock In {\em ECCV}, pages 740--755. Springer, 2014.

\bibitem{q2l}
Shilong Liu, Lei Zhang, Xiao Yang, Hang Su, and Jun Zhu.
\newblock Query2label: A simple transformer way to multi-label classification.
\newblock {\em arXiv preprint arXiv:2107.10834}, 2021.

\bibitem{REWC}
Xialei Liu, Marc Masana, Luis Herranz, Joost Van~de Weijer, Antonio~M Lopez,
  and Andrew~D Bagdanov.
\newblock Rotate your networks: Better weight consolidation and less
  catastrophic forgetting.
\newblock In {\em International Conference on Pattern Recognition}, pages
  2262--2268. IEEE, 2018.

\bibitem{IOD2023}
Yaoyao Liu, Bernt Schiele, Andrea Vedaldi, and Christian Rupprecht.
\newblock Continual detection transformer for incremental object detection.
\newblock In {\em CVPR}, pages 23799--23808, 2023.

\bibitem{GEM}
David Lopez-Paz et~al.
\newblock Gradient episodic memory for continual learning.
\newblock In {\em NIPS}, pages 6467--6476, 2017.

\bibitem{PackNet}
Arun Mallya and Svetlana Lazebnik.
\newblock Packnet: Adding multiple tasks to a single network by iterative
  pruning.
\newblock In {\em CVPR}, pages 7765--7773, 2018.

\bibitem{seg_il2022}
Andrea Maracani, Umberto Michieli, Marco Toldo, and Pietro Zanuttigh.
\newblock Recall: Replay-based continual learning in semantic segmentation.
\newblock In {\em ICCV}, pages 7026--7035, 2021.

\bibitem{iss-2021er2}
Andrea Maracani, Umberto Michieli, Marco Toldo, and Pietro Zanuttigh.
\newblock Recall: Replay-based continual learning in semantic segmentation.
\newblock {\em ICCV}, Jan 2021.

\bibitem{mccloskey1989catastrophic}
Michael McCloskey and Neal~J Cohen.
\newblock Catastrophic interference in connectionist networks: The sequential
  learning problem.
\newblock In {\em Psychology of Learning and Motivation}, volume~24, pages
  109--165. Elsevier, 1989.

\bibitem{SDR}
Umberto Michieli and Pietro Zanuttigh.
\newblock Continual semantic segmentation via repulsion-attraction of sparse
  and disentangled latent representations.
\newblock In {\em CVPR}, Jun 2021.

\bibitem{ICARL}
Sylvestre-Alvise Rebuffi, Alexander Kolesnikov, Georg Sperl, and Christoph~H
  Lampert.
\newblock icarl: Incremental classifier and representation learning.
\newblock In {\em CVPR}, pages 2001--2010, 2017.

\bibitem{asl2020}
Tal Ridnik, Emanuel Ben-Baruch, and et al.
\newblock Asymmetric loss for multi-label classification.
\newblock In {\em ICCV}, pages 82--91, 2021.

\bibitem{tresnet}
Tal Ridnik, Hussam Lawen, Asaf Noy, Emanuel Ben~Baruch, Gilad Sharir, and
  Itamar Friedman.
\newblock Tresnet: High performance gpu-dedicated architecture.
\newblock In {\em IEEE/CVF Winter Conference on Applications of Computer
  Vision}, pages 1400--1409, 2021.

\bibitem{ERbase}
Matthew Riemer, Ignacio Cases, and et al.
\newblock Learning to learn without forgetting by maximizing transfer and
  minimizing interference.
\newblock {\em arXiv preprint arXiv:1810.11910}, 2018.

\bibitem{PNN}
Andrei~A Rusu, Neil~C Rabinowitz, Guillaume Desjardins, Hubert Soyer, James
  Kirkpatrick, Koray Kavukcuoglu, Razvan Pascanu, and Raia Hadsell.
\newblock Progressive neural networks.
\newblock {\em arXiv preprint arXiv:1606.04671}, 2016.

\bibitem{oewc}
Jonathan Schwarz, Wojciech Czarnecki, and et al.
\newblock Progress \& compress: A scalable framework for continual learning.
\newblock In {\em ICML}, pages 4528--4537. PMLR, 2018.

\bibitem{hat}
Joan Serra, Didac Suris, Marius Miron, and et al.
\newblock Overcoming catastrophic forgetting with hard attention to the task.
\newblock In {\em ICML}, pages 4548--4557. PMLR, 2018.

\bibitem{CwD}
Yujun Shi, Kuangqi Zhou, Jian Liang, Zihang Jiang, Jiashi Feng, Philip~HS Torr,
  Song Bai, and Vincent~YF Tan.
\newblock Mimicking the oracle: An initial phase decorrelation approach for
  class incremental learning.
\newblock In {\em CVPR}, pages 16722--16731, 2022.

\bibitem{shim2021online}
Dongsub Shim, Zheda Mai, Jihwan Jeong, Scott Sanner, Hyunwoo Kim, and Jongseong
  Jang.
\newblock Online class-incremental continual learning with adversarial shapley
  value.
\newblock In {\em AAAI}, volume~35, pages 9630--9638, 2021.

\bibitem{iod2017}
Konstantin Shmelkov, Cordelia Schmid, and Karteek Alahari.
\newblock Incremental learning of object detectors without catastrophic
  forgetting.
\newblock In {\em ICCV}, pages 3400--3409, 2017.

\bibitem{sun2022information}
Shengyang Sun, Daniele Calandriello, Huiyi Hu, Ang Li, and Michalis Titsias.
\newblock Information-theoretic online memory selection for continual learning.
\newblock {\em arXiv preprint arXiv:2204.04763}, 2022.

\bibitem{tao2020topology}
Xiaoyu Tao, Xinyuan Chang, Xiaopeng Hong, Xing Wei, and Yihong Gong.
\newblock Topology-preserving class-incremental learning.
\newblock In {\em ECCV}, pages 254--270. Springer, 2020.

\bibitem{topic}
Xiaoyu Tao, Xiaopeng Hong, Xinyuan Chang, Songlin Dong, Xing Wei, and Yihong
  Gong.
\newblock Few-shot class-incremental learning.
\newblock In {\em CVPR}, pages 12183--12192, 2020.

\bibitem{gdtrs}
Hugo Touvron, Matthieu Cord, Alexandre Sablayrolles, Gabriel Synnaeve, and
  Herv{\'e} J{\'e}gou.
\newblock Going deeper with image transformers.
\newblock In {\em ICCV}, pages 32--42, 2021.

\bibitem{foster}
Fu-Yun Wang, Da-Wei Zhou, Han-Jia Ye, and De-Chuan Zhan.
\newblock Foster: Feature boosting and compression for class-incremental
  learning.
\newblock In {\em ECCV}, pages 398--414. Springer, 2022.

\bibitem{RMAM2017}
Zhouxia Wang, Tianshui Chen, Guanbin Li, Ruijia Xu, and et al.
\newblock Multi-label image recognition by recurrently discovering attentional
  regions.
\newblock In {\em ICCV}, pages 464--472, 2017.

\bibitem{l2p}
Zifeng Wang, Zizhao Zhang, and et al.
\newblock Learning to prompt for continual learning.
\newblock In {\em CVPR}, pages 139--149, 2022.

\bibitem{herding}
Max Welling.
\newblock Herding dynamical weights to learn.
\newblock In {\em ICML}, pages 1121--1128, 2009.

\bibitem{2020disloss}
Tong Wu, Qingqiu Huang, Ziwei Liu, Yu Wang, and Dahua Lin.
\newblock Distribution-balanced loss for multi-label classification in
  long-tailed datasets.
\newblock In {\em ECCV}, pages 162--178. Springer, 2020.

\bibitem{BIC}
Yue Wu, Yinpeng Chen, Lijuan Wang, Yuancheng Ye, Zicheng Liu, Yandong Guo, and
  Yun Fu.
\newblock Large scale incremental learning.
\newblock In {\em CVPR}, pages 374--382, 2019.

\bibitem{aanet}
Haofei Xu and Juyong Zhang.
\newblock Aanet: Adaptive aggregation network for efficient stereo matching.
\newblock In {\em CVPR}, pages 1959--1968, 2020.

\bibitem{der}
Shipeng Yan, Jiangwei Xie, and Xuming He.
\newblock Der: Dynamically expandable representation for class incremental
  learning.
\newblock In {\em CVPR}, pages 3014--3023, 2021.

\bibitem{yoon2021online}
Jaehong Yoon, Divyam Madaan, Eunho Yang, and Sung~Ju Hwang.
\newblock Online coreset selection for rehearsal-based continual learning.
\newblock {\em arXiv preprint arXiv:2106.01085}, 2021.

\bibitem{SI}
Friedemann Zenke, Ben Poole, and Surya Ganguli.
\newblock Continual learning through synaptic intelligence.
\newblock In {\em ICML}, pages 3987--3995. JMLR. org, 2017.

\bibitem{bic2020}
Bowen Zhao, Xi Xiao, Guojun Gan, Bin Zhang, and Shu-Tao Xia.
\newblock Maintaining discrimination and fairness in class incremental
  learning.
\newblock In {\em CVPR}, pages 13208--13217, 2020.

\end{thebibliography}
}
\clearpage
\begin{table*}[htbp]
\renewcommand\arraystretch{1.35}
\footnotesize
\centering
\begin{tabular}{l|c|cccccc}
\hline
\multirow{2}{*}{Method}& \multirow{2}{*}{Buffer Size} & \multicolumn{4}{c}{sessions} & \multicolumn{1}{c}{Average}  &Last mAP\% \\
  \cline{3-6}& & 1 & 2 & 3 & 4 & mAP\% &impro.\\
  \hline
FT~\cite{asl2020}&0 & 86.50 & 	54.6 & 26.53 &	27.88 &	51.38 & 47.37 \\ \midrule
iCaRL~\cite{ICARL}& \multirow{6}{*}{20/class}& 86.50 & 70.72 &	66.56 &	52.41 & 69.05 & 22.84\\
BIC ~\cite{BIC}& & 86.50 &	75.16  &	68.08 &	51.51 &	70.31 & 23.74 \\
ER~\cite{ERbase}& &86.50 &	69.48 &	60.74 &	58.04 &	68.69  & 17.21 \\
TPCIL~\cite{tao2020topology}&& 86.50 & 75.13 &67.81 &	63.41& 73.21& 11.84 \\
PODNet~\cite{podnet} &&86.50 &	75.71 &	70.52 &	66.96 &	74.92& 8.29 \\
DER++~\cite{der+} &&86.50& 75.84 &73.28 &	68.17&	75.95& 7.08 \\
\textbf{KRT-R}(Ours)& &\textbf{86.63} &	\textbf{78.84} &\textbf{77.29}&\textbf{75.25}& \textbf{79.47}&-\\\hline
\end{tabular}
\caption{Comparison results on MS-COCO dataset with B0-C20 benchmark}
\label{tb:b0c20}
\end{table*}

\begin{table*}[htbp]
\renewcommand\arraystretch{1.35}
\footnotesize
\centering
\begin{tabular}{l|c|cccccccccc}
\hline
\multirow{2}{*}{Method}& \multirow{2}{*}{Buffer Size} & \multicolumn{8}{c}{sessions} & \multicolumn{1}{c}{Average}  &Last mAP\% \\
  \cline{3-10}& & 1 & 2 & 3 & 4 & 5 & 6 &7 & 8 & mAP\% &impro.\\
  \hline
FT~\cite{asl2020}&0 & 92.51 &59.68 &41.23&	31.86&	22.03&	23.07&	19.27&	16.93&38.33& 53.24\\ \midrule
iCaRL~\cite{ICARL}& \multirow{6}{*}{20/class}& 92.51&73.62&	67.38&	60.20&	52.19&	43.63&44.25&	43.84&59.72& 26.33\\
BIC ~\cite{BIC}& &92.51&79.53&73.97&64.16&57.17&49.79&50.63&50.95 &64.96 & 19.22\\
ER~\cite{ERbase}& &92.51&75.14&	61.34&	56.83&	48.55&	51.44&	49.28&	47.19&60.28 & 22.98\\
TPCIL~\cite{tao2020topology}&& 92.51 &77.86	&69.12&	67.34&	62.85	&63.25&	62.12&	60.57&	69.45& 9.60\\
PODNet~\cite{podnet} &&92.51&79.95&	73.45&	68.22&	63.17&	62.98&	60.36&	58.82&	69.93& 11.35\\
DER++~\cite{der+} &&92.51&79.23&76.27 &70.68&68.88&	67.12&	64.16&	63.11&	72.74& 7.06\\
\textbf{KRT-R}(Ours)& &\textbf{92.25} &	\textbf{81.30} &\textbf{77.26}&\textbf{74.69}& \textbf{73.22}&\textbf{72.80}&\textbf{70.61}&\textbf{70.17}&\textbf{76.54}&-\\\hline
\end{tabular}
\caption{Comparison results on MS-COCO dataset with B0-C10 benchmark.}
\label{tb:b0c10}
\end{table*}

\begin{table*}[htbp]
\renewcommand\arraystretch{1.35}
\footnotesize
\centering
\begin{tabular}{l|c|ccccccc}
\hline
\multirow{2}{*}{Method}& \multirow{2}{*}{Buffer Size} & \multicolumn{5}{c}{sessions} & \multicolumn{1}{c}{Average}  &Last mAP\% \\
  \cline{3-7}& & 1 & 2 & 3 & 4 &5& mAP\% &impro.\\
  \hline
FT~\cite{asl2020}&0 & 82.88	&28.55&	28.06&	18.78&	16.99&	35.05 &58.19\\ \midrule
iCaRL~\cite{ICARL}& \multirow{6}{*}{20/class}& 82.41&67.87&	63.40&	58.27	&55.74	&65.61 & 19.44\\
BIC ~\cite{BIC}& & 82.41&71.51&	61.76&	55.52	&55.91&	65.55& 19.27\\
ER~\cite{ERbase}& & 82.41&	70.79&	66.58	&63.34	&61.59&	68.94 &13.59\\
TPCIL~\cite{tao2020topology}&& 82.41&	72.62&	71.65&	68.62&	66.54&	72.37&8.64\\
PODNet~\cite{podnet} &&82.41&	71.40&	70.76&	65.90&	64.22&	70.96&10.96\\
DER++~\cite{der+} &&82.41&	77.07&	73.92&	68.11&	66.31&	73.56&8.87\\
\textbf{KRT-R}(Ours)& &\textbf{82.37} &	\textbf{79.54} &\textbf{78.27}&\textbf{75.95}& \textbf{75.18}&\textbf{78.26}&-\\\hline
\end{tabular}
\caption{Comparison results on MS-COCO dataset with B40-C10 benchmark.}
\label{tb:b40c10}
\end{table*}

\clearpage
\begin{table*}[htbp]
\renewcommand\arraystretch{1.35}
\footnotesize
\centering
\begin{tabular}{l|c|ccccccccccc}
\hline
\multirow{2}{*}{Method}& \multirow{2}{*}{Buffer Size} & \multicolumn{9}{c}{sessions} & \multicolumn{1}{c}{Average}  &Last mAP\% \\
  \cline{3-11}& & 1 & 2 & 3 & 4 & 5 & 6 &7 & 8 & 9& mAP\% &impro.\\
  \hline
FT~\cite{asl2020}&0 & 82.41&45.74&16.43&17.59&	13.6&12.51&	10.99&	10.44&	10.66&24.49&61.77\\ \midrule
iCaRL~\cite{ICARL}& \multirow{6}{*}{20/class}& 82.41&74.1&62.8&60.4&61.6&56.4&54.9&	54.8&53.9&62.37&18.53 \\
BIC ~\cite{BIC}& &82.41&75.50&61.35&56.13&57.37&52.94&52.17&51.97&51.72&60.17&20.71\\
ER~\cite{ERbase}& &82.41&74.24&	66.58&62.92&63.88&61.03&60.30&59.96&58.12&65.49&14.31\\
TPCIL~\cite{tao2020topology}&& 82.41&74.91&	70.21&69.04&68.35&66.25&66.01&64.79&64.24&69.58&8.19\\
PODNet~\cite{podnet} &&82.41&76.18&68.21&66.31&66.45&61.45&61.5&59.98&58.89&66.82&13.54\\
DER++~\cite{der+} &&82.41&78.2&74.15&70.09&67.03&61.9&61.91&62.97&62.14&68.98&10.29\\
\textbf{KRT-R}(Ours)& &\textbf{82.37} &	\textbf{80.63} &\textbf{78.15}&\textbf{76.46}& \textbf{76.43}&\textbf{74.34}&\textbf{73.62}&\textbf{72.82}&\textbf{72.43}&\textbf{76.36}&-\\\hline
\end{tabular}
\caption{Comparison results on MS-COCO dataset with B40-C5 benchmark.}
\label{tb:b40c5}
\end{table*}

\begin{table*}[htbp]
\renewcommand\arraystretch{1.35}
\footnotesize
\centering
\begin{tabular}{l|c|ccccccc}
\hline
\multirow{2}{*}{Method}& \multirow{2}{*}{Buffer Size} & \multicolumn{5}{c}{sessions} & \multicolumn{1}{c}{Average}  &Last mAP\% \\
  \cline{3-7}& & 1 & 2 & 3 & 4 &5& mAP\% &impro.\\
  \hline
FT~\cite{asl2020}&0 & 99.25&94.16&85.51&68.67&62.88&82.09 &20.55\\ \midrule
iCaRL~\cite{ICARL}& \multirow{6}{*}{2/class}& 99.25&95.03&88.31&81.16&72.38&87.23&11.05 \\
BIC ~\cite{BIC}& &99.25&94.48&86.29&81.83&72.24&86.82&11.19 \\
ER~\cite{ERbase}& & 99.25&95.05&89.22&75.6&71.49&86.12 &11.94\\
TPCIL~\cite{tao2020topology}&& 99.25&95.1&88.04&78.15&77.35&87.58&6.08\\
PODNet~\cite{podnet} &&99.25&95.64&88.71&	80.28&76.60&88.09&6.83\\
DER++~\cite{der+} &&99.25&95.35&89.02&80.13&76.05&87.96&7.38\\
\textbf{KRT-R}(Ours)& &\textbf{99.77} &	\textbf{96.06} &\textbf{89.06}&\textbf{85.43}& \textbf{83.43}&\textbf{90.73}&-\\\hline
\end{tabular}
\caption{Comparison results on PASCAL VOC dataset with B0-C4 benchmark.}
\label{tb:b0c4}
\end{table*}

\begin{table*}[htbp]
\renewcommand\arraystretch{1.35}
\footnotesize
\centering
\begin{tabular}{l|c|cccccccc}
\hline
\multirow{2}{*}{Method}& \multirow{2}{*}{Buffer Size} & \multicolumn{6}{c}{sessions} & \multicolumn{1}{c}{Average}  &Last mAP\% \\
  \cline{3-8}& & 1 & 2 & 3 & 4 &5 & 6 & mAP\% &impro.\\
  \hline
FT~\cite{asl2020}&0 & 97.09&	86.97&	82.49&	61.65&	49.54&43.01&70.12&37.45\\ \midrule
iCaRL~\cite{ICARL}& \multirow{6}{*}{2/class}&97.09&89.32&84.37&69.74&66.63&66.70&78.98&13.76 \\
BIC ~\cite{BIC}&&97.09&89.94&85.33&76.03&72.04&69.71&81.69&10.75 \\
ER~\cite{ERbase}& & 97.09&89.33&87.74&76.86&69.26&68.58&81.48 &11.88\\
TPCIL~\cite{tao2020topology}&& 97.09&88.19&82.79&74.44&70.69&70.77&80.66&9.69\\
PODNet~\cite{podnet} &&97.09&89.89&85.59&72.13&71.18&71.35&81.21&9.11\\
DER++~\cite{der+} &&97.09&89.44&87.72&76.36&72.97&70.64&82.37&9.82\\
\textbf{KRT-R}(Ours)& &\textbf{97.64} &	\textbf{93.52} &\textbf{88.42}&\textbf{85.10}& \textbf{81.28}&\textbf{80.46}&\textbf{87.73}&-\\\hline
\end{tabular}
\caption{Comparison results on PASCAL VOC dataset with B10-C2 benchmark.}
\label{tb:b10c2}
\end{table*}

\end{document}